\pdfoutput=1
\documentclass[11pt]{article}

\usepackage[final]{acl}

\usepackage{times}
\usepackage{latexsym}

\usepackage[T1]{fontenc}
\usepackage[utf8]{inputenc}

\usepackage{microtype}

\usepackage{inconsolata}

\usepackage{graphicx}
\usepackage{dsfont}
\usepackage{amsmath}
\usepackage{amssymb}
\usepackage[english]{babel}
\usepackage{amsthm}
\usepackage{multirow}
\usepackage{makecell}
\usepackage{graphicx}
\usepackage{verbatim}
\usepackage{booktabs}
\usepackage{arydshln}
\usepackage{xcolor}
\usepackage{paralist}
\usepackage{colortbl}
\usepackage{cleveref}
\usepackage{transparent}
\usepackage{nicefrac}
\usepackage{subcaption}
\usepackage{enumitem}
\newcolumntype{C}[1]{>{\centering\let\newline\\\arraybackslash\hspace{0pt}}m{#1}}
\newcolumntype{R}[1]{>{\PreserveBackslash\raggedleft}p{#1}}
\newcolumntype{L}[1]{>{\PreserveBackslash\raggedright}p{#1}}

\title{\texttt{DRPruning}: Efficient Large Language Model Pruning through Distributionally Robust Optimization}

\author{
  Hexuan Deng\textsuperscript{\rm 1}~~~
  Wenxiang Jiao~~~
  Xuebo Liu\textsuperscript{\rm 1}\thanks{~~Corresponding Author}~~~
  Jing Li\textsuperscript{\rm 1}~~~
  Min Zhang\textsuperscript{\rm 1}~~~
  Zhaopeng Tu
  \\
  \textsuperscript{\rm 1}Institute of Computing and Intelligence, Harbin Institute of Technology, Shenzhen, China
  \\
  \texttt{\{hxuandeng,wenxiangjiaonju,tuzhaopeng\}@gmail.com},
  \\
  \texttt{\{liuxuebo,li.jing,zhangmin2021\}@hit.edu.cn}
}

\begin{document}
\maketitle
\begin{abstract}
Large language models (LLMs) deliver impressive results but face challenges from increasing model sizes and computational costs. Structured pruning reduces model size and speeds up inference but often causes uneven degradation across domains, leading to biased performance. 
To address this, we propose \textit{DRPruning}, a method that dynamically adjusts the data distribution during training to restore balanced performance across heterogeneous and multi-tasking data. 
Experiments in monolingual and multilingual settings show that DRPruning surpasses similarly sized models in both pruning and continued pretraining over perplexity, downstream tasks, and instruction tuning. 
Further analysis demonstrates the robustness of DRPruning towards various domains and distribution shifts. 
Furthermore, DRPruning can determine optimal reference losses and data ratios automatically, suggesting potential for broader applications. 
Code and scripts are available at~\url{https://github.com/hexuandeng/DRPruning}.
\end{abstract}

\section{Introduction}

Large language models (LLMs) have advanced rapidly, achieving impressive results across a wide range of tasks~\citep{ChatGPTGoodTranslator_2023, MathematicalCapabilitiesChatGPT_2023, ImprovingSimultaneous_DDL+23, ChatGPTKnowledgeableInexperienced_2023, commonIT}. 
However, this progress has come with increasing model sizes, significantly raising computational costs for both training and inference, which impacts their accessibility. Structured pruning is a promising approach to reduce model size~\citep{DeepCompression_HMD16, LearningStructured_WWW+16}, but it often causes uneven performance degradation across domains, leading to biased capabilities and unfair downstream task performance~\citep{ShearedLLaMA_XGZC23}.

Given that LLMs inherently handle heterogeneous, multi-domain data, distribution robustness becomes essential. A commonly used approach is distributionally robust optimization (DRO; \citealp{DistributionallyRobust_OSHL19, DistributionallyRobust_SKHL20}), which aims to optimize worst-case performance across distributions. 
A reference loss is defined for each domain as a target. Domains with larger deviations from this reference loss are assigned higher weights, while not straying too far from a predefined reference data ratio. 
However, setting these hyperparameters is challenging, and suboptimal configurations often result in poor outcomes~\citep{DistributionallyRobust_ZLL+21}.

To address this, we propose DRPruning, a distributionally robust pruning method that incorporates DRO to dynamically adjust the data distribution during training.
Further, using scaling laws~\citep{ScalingLaws_KMH+20, ScalingLaws_GFF+21}, we predict the loss after training as the reference loss, where larger deviations indicate poorer performance, thereby promoting capability recovery in these areas. 
Additionally, we gradually increase the reference data ratio for domains with greater deviations, ensuring robustness across a wider range of distributions, particularly more challenging ones.

DRPruning is validated through experiments in monolingual and multilingual settings, which represent varying degrees of distributional shift. 
DRPruning outperforms other data scheduling methods in both pruning and continued pretraining, as measured by perplexity (-5.59\%), downstream tasks (+1.52\%), and instruction tuning (55.4\% win rate).
Particularly in multilingual settings, DRPruning achieves +2.95\% in downstream tasks. 
To further assess domain-specific performance, we develop a sentence continuation benchmark using existing unlabeled data, demonstrating our improved domain-level capabilities (+17.9\%). 

Our contributions are summarized as follows:
\begin{itemize}%[leftmargin=10pt]
    \item DRPruning tackles domain imbalance in structured pruning by introducing a distributionally robust pruning method, that dynamically adjusts data ratios during training to ensure robustness against distributional shifts.
    \item We validate DRPruning through extensive experiments in monolingual and multilingual settings. Further analysis confirms its advantages in handling data heterogeneity and distribution shifts.
    \item DRPruning offers refined reference losses and data ratios, which can be applied more broadly to enhance various model training processes and contribute to advancements for LLMs.
\end{itemize}

\section{Background}

\subsection{Structured Pruning}
\label{sec:pruning}
To prune the model to any target configuration, we adopt structured pruning based on Sheared Llama \citep{ShearedLLaMA_XGZC23}. For each granularity $i$, pruning masks $Z = \{\mathbf{z}^i \mid \mathbf{z}^i \in \mathbb{R}^{D_i} \}$ are learned to determine whether substructures are pruned or retained, where $z^i_j = 0$ indicates pruning of the $j$-th substructure. Pruning is applied at various granularities, including transformer layers, hidden dimensions, attention heads, and FFN intermediate dimensions.

To parameterize the masks, the $\ell_0$ regularization method~\citep{LearningSparse_LWK18} with hard concrete distributions is used to concentrate probability mass at 0 or 1. Lagrange multipliers are then used to ensure the pruned model meets the target configuration. Specifically, if exactly \( t^i \) parameters must be retained for \( \mathbf{z}^i \), the following constraint is imposed:
\begin{equation}
\begin{aligned}
\noalign{\vskip-10pt} & \tilde{\ell}^i = \lambda^i \left(\sum_j z^i_j - t^i \right) + \phi^i \left(\sum_j z^i_j - t^i \right)^2. \\[-8pt]
\end{aligned}
\end{equation}

The final training loss integrates these constraints with the language modeling loss of the pruned model, jointly optimizing the model parameters \(\theta\) and pruning masks \(\mathbf{z}\), with \(\mathbf{z}\) typically uses a higher learning rate. After pruning, the highest-scoring components are retained.

\subsection{Distributionally Robust Optimization}
\label{sec:distRobust}

To mitigate uneven domain performance after pruning, we apply distributionally robust optimization (DRO; \citealp{DistributionallyRobust_OSHL19, DistributionallyRobust_SKHL20}) to improve the model's robustness to distribution shifts. DRO seeks a model $\theta$ that performs well across a set of potential test distributions $\mathcal{Q}$ over $n$ domains. Formally:
\begin{equation}
\begin{aligned}
\noalign{\vskip-4pt} \underset{\theta}{\operatorname{minimize}} \sup _{Q \in \mathcal{Q}} \mathbb{E}_{(\mathbf{x}, \mathbf{y}) \sim Q}[\ell(\mathbf{x}, \mathbf{y} ; \theta)]. \\[-4pt]
\end{aligned}
\label{eq:minmax}
\end{equation}

To solve the min-max optimization, the iterative best response algorithm \citep{fudenberg1998theory} is used. Each iteration consists of first performing the empirical risk minimization on the current data distribution $\mathbf{q}^t$, followed by updating the data distribution using worst-case weights based on the current parameters. Formally,
\begin{equation}
\begin{aligned}
\noalign{\vskip-5pt} & \theta^{t+1} \leftarrow \underset{\theta}{\operatorname{argmin}} \sum_i q_i^t \ell\left(\theta ; D_i\right), \\[-3pt]
& \mathbf{q}^{t+1} \leftarrow \underset{\mathbf{q} = \{q_1, \ldots, q_n\} \in \mathcal{Q}}{\operatorname{argmax}} \sum_i q_i \ell\left(\theta^{t+1} ; D_i\right). \\
\end{aligned}
\label{eq:naivedro}
\end{equation}

\section{Our Proposed DRPruning Method}

\begin{figure*}[!ht]
    \centering
    \includegraphics[width=0.95\linewidth]{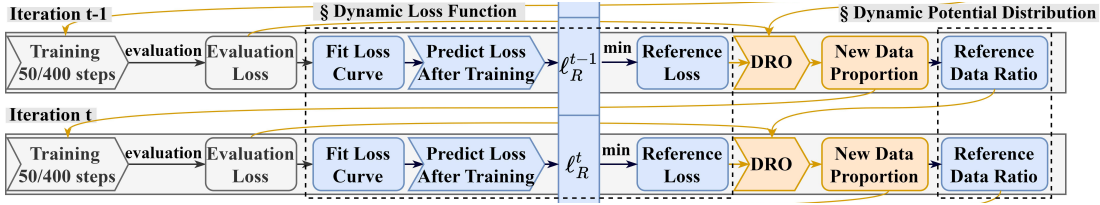}
    \caption{\label{fig:main}
    Data proportion update procedure for DRPruning. The gray part represents the standard training process, the yellow part represents the normal process for DRO, and the blue part represents our newly added module.}
\end{figure*}

To address the challenges of LLMs in handling heterogeneous and multi-tasking data, we propose DRPruning, illustrated in Figure~\ref{fig:main}. Each evaluation phase is treated as an iteration: the evaluation loss first updates the \textbf{reference loss}, which, along with the previous \textbf{reference data ratio}, serves as input to the DRO process. This yields a new data proportion for the next training step, and the reference data ratio is updated accordingly.

\subsection{Distributionally Robust Pruning}
\label{sec:DRPruning}

We first introduce the overall procedure by applying na\"ive DRO to design an effective pruning and continued pretraining method. Specifically, we adopt common techniques \citep{LLMPrunerStructural_MFW23, NutePruneEfficient_LCHB24}, first applying structured pruning to reduce model parameters, followed by continued pretraining to restore capabilities. Compared to training from scratch, this approach requires fewer unlabeled data to restore the model's performance \citep{LoRAPruneStructured_ZCS+24}.

\paragraph{Integrate DRO into pruning and continued pretraining.} During training, we use DRO to dynamically adjust the data ratio to improve the model's robustness and convergence speed. Specifically, to prevent overfitting, we compile a validation set and use the evaluation loss as the loss score. After each evaluation, we update the data ratio based on the evaluation loss using the DRO method, as in Eqn.~\ref{eq:naivedro}, guiding training to focus more on underperforming domains.

\paragraph{Further improvement.} Next, we optimize the loss function $\ell$ (Section~\ref{sec:refLoss}) and potential distributions $\mathcal{Q}$ (Section~\ref{sec:refRatio}) to ensure robust training, as shown in Figure~\ref{fig:main}. In contrast, Sheared Llama employs a dynamic scheduling strategy that forces the model to strictly adhere to the relative loss magnitudes of larger LLMs, without placing any constraints on the potential distributions. This leads to suboptimal results, particularly in multilingual settings with significant distribution shifts.

\subsection{Dynamic Loss Function}
\label{sec:refLoss}

To stabilize DRO training and prevent domains with slow convergence from disproportionately influencing the weights, the use of a \textit{\textbf{reference loss}} $\ell_R$ is a common approach \citep{DistributionallyRobust_OSHL19, DistributionallyRobust_ZLL+21}. This reference loss establishes the minimum acceptable performance for a domain. Furthermore, we update the loss score as $\ell\left(\theta; D\right) \leftarrow \ell\left(\theta; D\right) - \ell_R$. Proper tuning of $\ell_R$ can significantly improve performance \citep{TencentsMultilingual_JTL+22}. However, determining an appropriate value remains a challenging task.

\paragraph{Minimum performance estimation.}
To address this, we predict the model's loss at the end of training as an estimate of the minimum acceptable performance. Specifically, we leverage scaling laws to capture training dynamics and forecast the loss based on evaluation loss trends \citep{ScalingLaws_KMH+20, WhenScaling_ZLCF24}. Given the number of parameters \(P\) and the current training step \(T\), the predicted training loss is estimated by:
\begin{equation}
\begin{aligned}
\noalign{\vskip-12pt} \hat{\ell}(P, T) = A \cdot \frac{1}{P^\alpha} \cdot \frac{1}{T^\beta} + E, \\[-12pt]
\end{aligned}
\end{equation}
where \(A\), \(E\), \(\alpha\), and \(\beta\) are trainable parameters. For each domain, after each evaluation, we collect a data point, refit the curve to all collected points, and use the predicted curve to estimate the loss at the end of training as the predicted minimum performance. Following \citet{TrainingComputeOptimal_HBM+22a}, we estimate using the Huber loss ($\delta$ = 0.001) and the L-BFGS algorithm, and select the average of the best-fitting three from a grid of initializations. To ensure sufficient data points, we start predictions only after 20\% of training is complete.

\paragraph{Reference loss adjustment.}
Subsequently, we set the reference loss using the predicted minimum performance. In our preliminary experiments, this approach exhibits strong numerical stability. To accelerate convergence, we adopt the minimum value as the reference loss. This dynamically evaluates domains with poorer performance, allowing DRO to assign higher weights to these domains, thereby promoting faster model convergence.

\subsection{Dynamic Potential Distribution}
\label{sec:refRatio}

\citet{DistributionallyRobust_SKHL20} consider robustness to arbitrary subpopulations, which is overly conservative and degenerates into training only on the highest-loss domain. To address this issue, \citet{DistributionallyRobust_ZLL+21} propose a more reasonable assumption by restricting $\mathcal{Q}$ in Eqn.~\ref{eq:minmax} to an $f$-divergence ball \citep{Csiszar67} around a \textit{\textbf{reference data ratio}} $\mathbf{p}_R$. This yields promising results, better ensuring domain balance \citep{TencentsMultilingual_JTL+22}. Formally,
\begin{equation}
\begin{aligned}
\noalign{\vskip-20pt} \mathcal{Q} = \left\{\mathbf{q}: \chi^2\left(\mathbf{q}, \mathbf{p}_R\right) \leq \rho \right\}. \\[-20pt]
\end{aligned}
\label{eq:x2ratio}
\end{equation}

However, this assumption can be too restrictive, necessitating a carefully chosen reference data ratio $\mathbf{p}_R$. An unreasonable choice may reduce the model's robustness to distributional shifts.

\paragraph{Reference data ratio adjustment.}
To address this, we propose a method that combines the strengths of the aforementioned approaches. We still employ Eqn.~\ref{eq:x2ratio} to constrain the distribution within a limited range, while gradually shifting the reference data ratio towards domains with higher losses to improve the model's robustness to more challenging distributions. To ensure adequate training across all traversed potential distributions, we gradually update the reference ratio.

Compared to existing reference ratios, the DRO method dynamically assigns higher weights $\mathbf{q}$ to domains with higher losses. This method shows good numerical stability, which we leverage to update the reference ratio. Formally, we update:
\begin{equation}
\begin{aligned}
\noalign{\vskip-8pt} \mathbf{p}^{t+1}_R = \delta \cdot \mathbf{q}^{t} + (1-\delta) \cdot \mathbf{p}^{t}_R. \\[-8pt]
\end{aligned}
\end{equation}

Finally, to prevent the method from degenerating into training solely on the highest-loss domain, we constrain the reference ratio of each domain to lie between $\frac{1}{n}$ and $n$ times the initial ratio. Formally, we set $\frac{1}{n} \cdot \mathbf{p}^0_R \leq \mathbf{p}^t_R \leq n \cdot \mathbf{p}^0_R$. We apply this method after 40\% of the training is completed, ensuring the model sufficiently converges near the initial reference ratio and that the reference loss stabilizes.

\section{Experiments}

\subsection{Experimental Setup}
\label{sec:setup}

\paragraph{Model.} Llama2-7B model~\citep{LlamaOpen_TMS+23} is used as the base model. We employ the same target architecture as Sheared Llama for structured pruning to ensure a fair comparison. We compare our method, i.e., \textbf{DRPruning}, to strong open-source models of similar sizes, including \textbf{Pythia}-1.4B and 2.8B~\citep{PythiaSuite_BSA+23} and \textbf{Sheared Llama}-1.3B and 2.7B. Additionally, we reproduce Sheared Llama, using the same data settings to control for other variables (\textbf{ReSheared}). Further details are provided in Appendix~\ref{apx:settingMain}.

\paragraph{Data.} To ensure comparability with Sheared Llama, we align most of our settings with its approach. However, due to insufficient documentation of its data filtering method, we are unable to replicate the results under the 2.7B setting. Therefore, our comparison primarily focuses on ReSheared and DRPruning, using a similar data setting for our reproduction. We allocate 0.4 billion tokens for pruning, utilizing the publicly available pruning dataset of Sheared Llama. We employ 50 billion tokens for continued pretraining, and use SlimPajama~\citep{SlimPajamaDCUnderstanding_STM+24}, a filtered version of RedPajama~\citep{together2023redpajama}, and use its training split for continued pretraining.

\paragraph{Downstream task evaluation.} We use the lm-evaluation-harness package~\citep{eval-harness} to evaluate on an extensive suite of downstream tasks:
\begin{itemize}%[leftmargin=10pt]
    \item We follow Llama2 to report the 0-shot performance on PIQA~\citep{PIQAReasoning_BZB+19}, WinoGrande (WinoG, \citealp{WinoGrandeAdversarial_SBBC19}), ARC Easy (ARCE, \citealp{ThinkYou_CCE+18}), SQuAD~\citep{KnowWhat_RJL18}, BoolQ~\citep{BoolQExploring_CLC+19}, TruthfulQA (TruthQA, \citealp{TruthfulQAMeasuring_LHE22}), and 5-shot performance on Natural Questions (NQ, \citealp{NaturalQuestions_KPR+19}) and TriviaQA (TriQA, \citealp{TriviaQALarge_JCWZ17}).
    \item We follow Pythia to report the 0-shot performance of LAMBADA (LAMB, \citealp{LAMBADADataset_PKL+16}), LogiQA~\citep{LogiQAChallenge_LCL+20}, SciQ~\citep{CrowdsourcingMultiple_WLG17}, and WSC~\citep{ReviewWinograd_KLD+20}.
    \item We follow Sheared Llama to report performance of the tasks used by Open LLM Leaderboard, including 10-shot HellaSwag (HelS, \citealp{HellaSwagCan_ZHB+19}), 25-shot ARC Challenge (ARCC, \citealp{ThinkYou_CCE+18}), and 5-shot MMLU~\citep{MeasuringMassive_HBB+21}.
\end{itemize}

\paragraph{Instruction tuning evaluation.}
To further explore the potential applications of the base model, we follow Sheared Llama by training with 10k instruction-response pairs sampled from the ShareGPT dataset and using another 1k instructions for evaluation. We follow \citet{LargeLanguage_WLC+24} and Sheared Llama to employ LLMs, specifically \texttt{GPT-4o}, as an evaluator to compare the responses of the two models and report the win rates. 

\begin{table}[!t]
\small
\centering
\scalebox{0.9}{
\begin{tabular}{lccccccccc}
\toprule
\bf Method & \bf From & \bf To & \bf PPL $\downarrow$ & \bf Task $\uparrow$ \\ \midrule
\bf Sheared Llama & 7B & 1.3B & 10.05 & 34.89 \\
\bf ReSheared & 7B & 1.3B & 10.42 & 34.85 \\
\bf DRPruning & 7B & 1.3B & \bf 9.83 & \bf 35.60 \\ \midrule
\bf Sheared Llama & 7B & 2.7B & 7.64 & 39.75 \\
\bf ReSheared & 7B & 2.7B & 7.83 & 39.98 \\
\bf DRPruning & 7B & 2.7B & \bf 7.40 & \bf 40.18 \\ \bottomrule
\end{tabular}}
\caption{\label{tab:prune_ppl}
Perplexity (PPL) and downstream task performance (Task) of pruned models. The ``Task'' performance represents the macro-average across 15 tasks, as detailed in Section~\ref{sec:setup}. ``From'' and ``To'' indicate the model size before and after pruning, respectively.}
\end{table}
\begin{figure}[!t]
    \centering
    \includegraphics[width=0.95\linewidth]{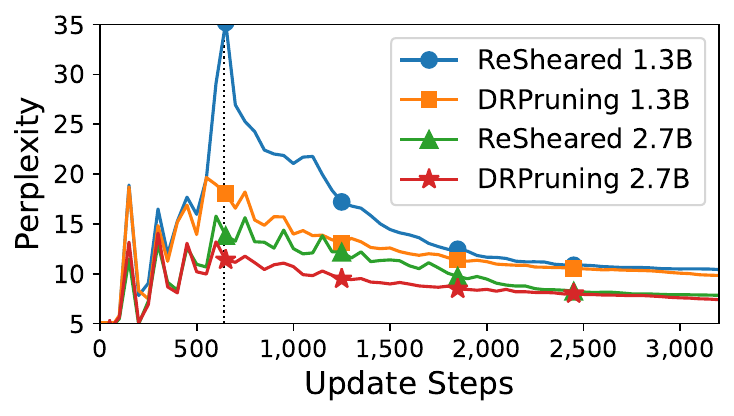}
    \caption{\label{fig:prune_ppl}
    The curve of PPL changes during pruning from 7B. Over the first 640 iterations (the vertical dash line), the model size is gradually reduced from 7B to the target size, which causes an initial increase in PPL.}
\end{figure}
\begin{table*}[!ht]
\small
\centering
\scalebox{0.9}{
\begin{tabular}{lC{1.25cm}C{1.2cm}C{1.26cm}C{1.48cm}C{1.3cm}C{1.2cm}C{1.26cm}C{1.48cm}C{1.3cm}}
\toprule
\multirow{2}[3]{*}{\bf Tasks} & \bf 7B & \multicolumn{4}{c}{\bf 2.7B} & \multicolumn{4}{c}{\bf 1.3B} \\ \cmidrule(lr){2-2}\cmidrule(lr){3-6}\cmidrule(lr){7-10}
~ & \bf Llama2$^\dag$ & \bf Pythia$^\dag$ & \bf Sheared$^\dag$ & \bf ReSheared & \bf DRPrun. & \bf Pythia$^\dag$ & \bf Sheared$^\dag$ & \bf ReSheared & \bf DRPrun. \\ \midrule
\bf WSC & 36.54 & 38.46 & \bf 48.08 & 36.54 & \underline{46.15} & 36.54 & 36.54 & \underline{40.38} & \bf 50.00 \\
\bf TriQA (5) & 64.16 & 27.17 & \underline{42.92} & 40.14 & \bf 43.33 & 18.19 & \underline{26.03} & 24.98 & \bf 28.10 \\
\bf NQ (5) & 25.98 & 7.12 & \underline{14.85} & 13.49 & \bf 15.82 & 4.79 & \underline{8.75} & 8.39 & \bf 10.44 \\
\bf TruthQA & 32.09 & 28.79 & \bf 30.21 & 28.41 & \underline{30.13} & \bf 30.75 & 29.12 & 28.09 & \underline{29.68} \\
\bf LogiQA & 30.11 & 28.11 & \underline{28.26} & 26.27 & \bf 28.73 & 27.50 & 27.50 & \underline{28.11} & \bf 28.88 \\
\bf BoolQ & 77.71 & 64.50 & \bf 65.99 & 64.92 & \underline{65.08} & \underline{63.30} & 62.05 & 61.01 & \bf 63.36 \\
\bf LAMB & 73.90 & 64.76 & \bf 68.21 & 66.18 & \underline{66.91} & \bf 61.67 & \underline{61.09} & 58.84 & 60.28 \\
\bf MMLU (5) & 44.18 & \bf 27.09 & 26.63 & 25.70 & \underline{26.99} & \underline{26.75} & 25.70 & 26.60 & \bf 27.28 \\
\bf SciQ & 94.00 & 88.50 & \bf 91.10 & 90.10 & 89.80 & 86.70 & \underline{87.00} & 86.40 & \bf 87.70 \\
\bf ARCE & 76.35 & 64.27 & \underline{67.34} & \bf 67.72 & 67.13 & 60.40 & \bf 60.90 & 60.35 & \bf 60.90 \\
\bf ARCC (25) & 52.65 & 36.35 & \bf 42.66 & 40.10 & \underline{40.53} & 33.02 & \underline{33.96} & \bf 34.30 & 33.62 \\
\bf PIQA & 78.07 & 73.88 & \underline{76.12} & \bf 76.71 & 75.19 & 70.84 & \underline{73.50} & \bf 74.59 & 72.69 \\
\bf WinoG & 69.06 & 59.83 & \bf 65.04 & 63.38 & \underline{64.72} & 57.38 & 57.85 & \bf 60.06 & \underline{58.01} \\
\bf SQuAD & 40.02 & 26.81 & \bf 49.26 & \underline{49.17} & 44.69 & 22.66 & 29.57 & \bf 37.59 & \underline{35.06} \\
\bf HelS (10) & 78.95 & 60.81 & \underline{71.24} & \bf 72.03 & 69.22 & 53.49 & \underline{61.05} & \bf 63.06 & 58.88 \\ \midrule
\bf Average & 58.25 & 46.43 & \bf 52.53 & 50.72 & \underline{51.63} & 43.60 & 45.37 & \underline{46.18} & \bf 46.99 \\
\bottomrule
\end{tabular}}
\caption{\label{tab:main}
Performances of different models across 15 downstream tasks. ``Sheared'' refers to Sheared Llama. ``ReSheared'' is our reproduction of Sheared Llama. ``DRPrun.'' refers to our method. The number of shots is indicated in parentheses, with 0-shot used when unspecified. A model marked with \dag\; indicates training on different data. \textbf{Bold} and \underline{underlined} represent the best and second-best results, respectively, for each model size.}
\end{table*}

\subsection{Main Results for Pruning}

Our method surpasses ReSheared during pruning in PPL and downstream tasks.

\paragraph{DRPruning promotes convergence by increasing the weight of underperformed domains.}
To demonstrate this, we record the average PPL across different domains on the validation set. Table~\ref{tab:prune_ppl} shows that our method achieves lower PPL.
Figure~\ref{fig:prune_ppl} confirms faster convergence as pruning proceeds, with further potential in later training stages, suggesting additional gains with extended training.

\paragraph{Our method better preserves the original model's performance during pruning.}\,\,For a comprehensive analysis, we evaluate the model's downstream task performance post-pruning. Across 15 tasks, our method significantly improves performance on average, with +1.53\% over the open-source model and +1.27\% in a fair comparison.

The pruning similarity analysis is detailed in Appendix~\ref{apx:analysisMask}, and Appendix~\ref{apx:largerLLM} shows that pruning larger LLMs offers no advantage.

\begin{table*}[!ht]
\small
\centering
\scalebox{0.9}{
\begin{tabular}{clcccccccc}
\toprule
& & \bf CC & \bf C4 & \bf GitHub & \bf Book & \bf Wiki & \bf ArXiv & \bf StackEx \\
\midrule
\multirow{4}*{\rotatebox{90}{\textbf{Data Ratio}}} & \bf Constant & 67.0\% & 15.0\% & 4.5\% & 4.5\% & 4.5\% & 2.5\% & 2.0\% \\
& \bf DRO & 54.6\% & 29.0\% & 3.7\% & 5.0\% & 3.8\% & 1.9\% & 1.9\% \\
& \bf DRPruning & 55.7\% & 15.6\% & 2.4\% & 3.7\% & 18.4\% & 2.1\% & 2.1\% \\ \cmidrule{2-9}
& \bf Reference Ratio vs. Constant & 47.1\% & 30.5\% & 2.9\% & 3.5\% & 9.1\% & 3.9\% & 3.0\% \\ \midrule
\multirow{2}*{\rotatebox{90}{\textbf{Loss}}} & \bf Evaluation Loss vs. Constant & +0.011 & -0.010 & +0.030 & +0.007 & -0.123 & +0.003 & +0.001 \\
& \bf Reference Loss vs. Constant & +0.067 & +0.162 & +0.093 & +0.094 & -0.108 & +0.014 & -0.053 \\ 
\bottomrule
\end{tabular}}
\caption{\label{tab:ablation}
Data usage ratios, hyperparameter adjustments, and domain-specific loss for Constant, na\"{i}ve DRO, and our strategy. The first three rows show the total data usage ratios during training, while the last three rows represent the hyperparameters and the evaluation loss at the end of training.}
\end{table*}

\subsection{Main Results for Continued Pretraining}

DRPruning outperforms ReSheared on average across LM benchmarks and instruction tuning.

\paragraph{LLMs recovered by our method are better foundation models after continued pretraining.}
Table~\ref{tab:main} presents the downstream performance of models with similar sizes, showing that our model surpasses most open-source models. We are unable to replicate the results of the 2.7B Sheared Llama due to differences in data, where our approach yields worse performance. Nevertheless, we outperform other open-source LLMs. On average, we achieve an improvement of +4.95\% comparing to open-source LLMs. Additionally, under consistent experimental conditions, our method outperforms ReSheared, achieving +1.78\% on average. Its statistical significance is confirmed by t-test in Appendix~\ref{apx:ttest}. Moreover, we achieve significantly lower perplexity: 5.61 vs 5.97 for 1.3B, and 5.00 vs 5.27 for 2.7B, averaging a -5.60\% reduction.

\begin{figure}[t]
    \centering
    \includegraphics[width=0.95\linewidth]{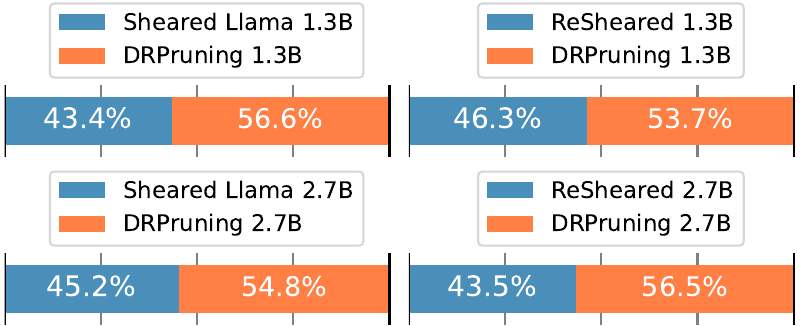}
    \caption{\label{fig:winrate}
    Win rate during instruction tuning. DRPruning outperforms Sheared Llama and ReSheared.}
\end{figure}

\begin{figure}[!t]
\begin{subfigure}{.24\textwidth}
  \centering
  \includegraphics[width=\textwidth]{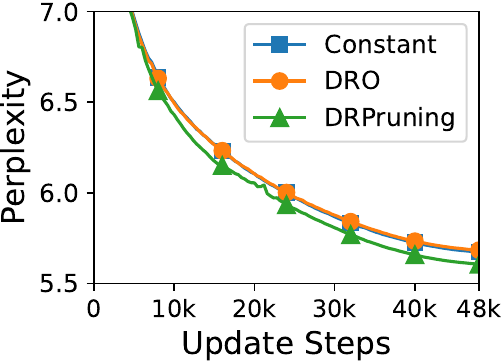}
\end{subfigure}%
\begin{subfigure}{.24\textwidth}
  \centering
  \includegraphics[width=\textwidth]{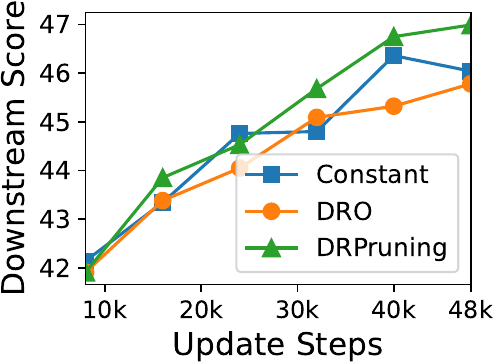}
\end{subfigure}
\caption{\label{fig:ablation}
Effectiveness of our method compared to constant scheduling and na\"{i}ve DRO during the 1.3B continued pretraining. 
Left figure: PPL trends; Right: average performance across 15 downstream tasks.}
\label{fig:test}
\end{figure}

\paragraph{The effectiveness of DRPruning is further demonstrated by instruction tuning.}
We compare the win rates of our instruction-tuned model with Sheared Llama and ReSheared. Figure~\ref{fig:winrate} shows our model achieves a 55.4\% win rate, surpassing both the open-source Sheared Llama and ReSheared. This highlights that DRPruning offers a stronger foundation for further use.

Additionally, DRPruning introduces no extra GPU computation, with the only overhead stemming from data ratio calculation, contributing to less than 1.5\% of the training time. Furthermore, we implemented parallel data ratio computation and demonstrated that it does not impact performance, as detailed in Appendix~\ref{apx:efficiency}.

\section{Analysis}

\subsection{Ablation Study}

We assess the impact of our dynamic data scheduling on 1.3B models, comparing it to constant scheduling and the na\"{i}ve DRO method.

\paragraph{DRPruning significantly outperforms DRO method.} As shown in Figure~\ref{fig:ablation}, our method consistently reduces PPL compared to the constant and DRO approach.
In downstream tasks, DRPruning also outperforms both baselines, with performance improving steadily in the mid to late stages of training (-0.26 for DRO, +0.95 for ours). This underscores the sensitivity of DRO to hyperparameters and the necessity of dynamic adjustments.

\paragraph{Our strategy dynamically identifies underperforming domains.}
To highlight how our method differs from DRO, Table \ref{tab:ablation} presents data usage ratios, hyperparameter adjustments, and domain-specific loss. Our strategy identifies Wiki as underperforming compared to optimal, assigning it a lower reference loss (-0.11) and a higher reference ratio (+4.6\%). This reduces loss on Wiki (-0.12) while keeping loss stable in other domains (maximum increase of +0.03). Besides, by dynamically adjusting the reference ratio, we improve the model's robustness across the full range of ratios from the initial to the new reference distribution. Compared to DRO, this offers robustness to a wider range of distribution shifts.

\begin{figure}[!t]
    \centering
    \includegraphics[width=0.95\linewidth]{pic/benchmark.pdf}
    \caption{\label{fig:benchmark} 
    The generation procedure for our sentence continuation task. The orange nodes represent data storage nodes, while the blue trapezoidal nodes represent data processing nodes.
    }
\end{figure}

\begin{table}[!t]
\small
\centering
\scalebox{0.9}{
\begin{tabular}{p{1cm}p{6.1cm}}
\toprule
\bf Question & If the latter described their efforts to adapt to European conditions,\\ \midrule
\bf Right Option & the former insisted that Muslims adhere to proper canons of learning and textual interpretation.\\ \midrule
\bf Wrong Option & it also highlighted the resilience and ingenuity that had brought them this far despite challenges.\\ \bottomrule
\end{tabular}}
\caption{\label{tab:case}
Case For the sentence continuation task.}
\end{table}

\begin{table*}[!ht]
\small
\centering
\scalebox{0.9}{
\begin{tabular}{clccccccccc}
\toprule
~ & \bf Model & \bf CC & \bf C4 & \bf GitHub & \bf Book & \bf Wiki & \bf ArXiv & \bf StackExchange & \bf Average \\
\midrule
\multirow{4}*{\rotatebox{90}{\small \textbf{1.3B}}} & \bf Constant & 35.00 & 40.00 & 88.00 & 47.75 & 21.75 & 82.00 & 72.50 & 55.29 \\
~ & \bf Sheared Llama & 21.75 & 26.00 & 93.75 & 33.50 & 29.50 & 38.50 & 56.50 & 42.79 \\
~ & \bf ReSheared & 30.50 & 30.25 & 89.25 & 32.00 & 23.00 & 81.00 & 47.50 & 47.64 \\
~ & \bf DRPruning & \bf 44.00 & \bf 51.50 & \bf 94.75 & \bf 48.00 & \bf 33.50 & \bf 86.50 & \bf 90.00 & \bf 64.04 \\
\midrule
\multirow{3}*{\rotatebox{90}{\small \textbf{2.7B}}} & \bf Sheared Llama & 81.25 & \bf 89.50 & 95.50 & \bf 96.50 & \bf 89.25 & 90.50 & 82.75 & \bf 89.32 \\
~ & \bf ReSheared & 61.75 & 60.25 & 96.00 & 73.00 & 80.50 & \bf 93.75 & \bf 92.00 & 79.61 \\
~ & \bf DRPruning & \bf 82.25 & 77.75 & \bf 99.00 & 86.75 & 87.50 & 79.50 & 89.25 & 86.00 \\
\bottomrule
\end{tabular}}
\caption{\label{tab:domain}
Domain-level results under the benchmark we generated. The abbreviations of tasks refer to the evaluation of seven domains used for training in RedPajama.
}
\end{table*}

\begin{table*}[!ht]
\small
\centering
\scalebox{0.9}{
\begin{tabular}{lcccccccccccc}
\toprule
\bf Base Model & \bf Prune & \bf PT & \bf Method & \bf EN & \bf RU & \bf ZH & \bf JA & \bf AR & \bf TR & \bf KO & \bf TH & \bf Average \\
\midrule
\bf XGLM-1.7B & X & X & \bf - & 55.06 & 52.97 & 51.02 & 51.00 & 42.89 & 37.99 & 49.00 & 38.63 & 47.32 \\
\bf Qwen1.5-1.8B & X & X & \bf - & 60.89 & 52.30 & \bf 56.13 & 53.30 & 42.17 & 34.98 & 48.25 & 36.75 & 48.10 \\
\bf Qwen2-1.5B & X & X & \bf - & 61.58 & 57.83 & 55.72 & 55.30 & 43.31 & 35.98 & 49.25 & 36.02 & 49.37 \\ \midrule
\bf Qwen2-1.5B & X & $\checkmark$ & \bf ReSheared & \bf 62.16 & 58.95 & 54.93 & \bf 55.60 & 43.91 & 37.27 & \bf 54.05 & 39.96 & 50.85 \\
\bf Qwen2-1.5B & X & $\checkmark$ & \bf DRPruning & 61.67 & \bf 59.09 & 54.01 & 54.95 & 45.14 & \bf 46.91 & 52.65 & \bf 44.42 & \bf 52.35 \\
\bf Qwen2-7B & $\checkmark$ & $\checkmark$ & \bf DRPruning & 60.43 & 56.80 & 55.72 & 55.05 & \bf 45.69 & 43.82 & 53.95 & 43.53 & 51.87 \\ \bottomrule
\end{tabular}}
\caption{\label{tab:multi}
The average performance on downstream tasks across multiple languages. ``Prune'' refers to the pruning procedure applied to the base model, while ``PT'' indicates continued pretraining on the provided dataset.
}
\end{table*}
\begin{table*}[!ht]
\small
\centering
\scalebox{0.9}{
\begin{tabular}{clcccccccc}
\toprule
& & \bf EN & \bf RU & \bf ZH & \bf JA & \bf AR & \bf TR & \bf KO & \bf TH \\
\midrule
\multirow{4}*{\rotatebox{90}{\textbf{Data Ratio}}} & \bf Default Reference Ratio & 27.7\% & 18.5\% & 13.0\% & 10.4\% & 9.1\% & 8.9\% & 6.7\% & 5.8\% \\
& \bf ReSheared & 82.8\% & 5.9\% & 5.5\% & 2.2\% & 1.0\% & 0.8\% & 1.2\% & 0.6\% \\
& \bf DRPruning & 19.0\% & 7.8\% & 12.9\% & 19.1\% & 9.2\% & 19.9\% & 6.7\% & 5.4\% \\ \cmidrule{2-10}
& \bf Reference Ratio vs. ReSheared & 23.1\% & 9.2\% & 8.5\% & 20.3\% & 8.6\% &  17.8\% & 6.7\% & 5.8\% \\ \midrule
\multirow{2}*{\rotatebox{90}{\textbf{Loss}}} & \bf Evaluation Loss vs. ReSheared & +0.143 & -0.067 & -0.123 & -0.330 & -0.465 & -0.841 & -0.282 & -0.304 \\
& \bf Reference Loss vs. ReSheared & +0.211 & +0.008 & -0.060 & -0.267 & -0.412 & -0.804 & -0.219 & -0.229 \\ \bottomrule
\end{tabular}}
\caption{\label{tab:multi_analysis}
Data usage ratios, hyperparameter adjustments, and domain-specific loss for reproduction of Sheared Llama (ReSheared) and our approach (DRPruning) during continued pretraining from Qwen2 1.5B.}
\end{table*}

\subsection{Robustness across Different Domains}

While our method shows clear advantages in PPL, a gap remains between PPL and downstream performance. The lack of domain-specific test sets also limits further analysis. To address this, we create downstream tasks from unlabeled datasets for detailed domain-specific evaluation.

\paragraph{Automatic construction of sentence continuation tasks across domains.}
To assess base model performance, we use sentence continuation tasks where the model selects the better continuation between two options. As shown in Figure~\ref{fig:benchmark}, we use the SlimPajama test set as the correct sentence and have the model generate the incorrect alternatives. First, ``LLM Filtering'' selects consecutive sentences with a causal relationship, then ``Split'' each sentence into two parts: the first as the question, and the second as the right option. For this large-scale filtering, we use \texttt{GPT-4o-mini}.

Next, ``LLM Rewrite'' generates incorrect options. Since LLMs struggle to create incorrect but related content, we follow \citet{NewTermBenchmarking_DJL+24} to first generate a reasonable continuation, then modify it to create an incorrect version. Finally, ``LLMs Verification'' scores both options and filters out cases where the scores don’t match the true answers. We use \texttt{GPT-4o} in these procedures. We further select questions with the largest score differences. In total, we select 400 questions per domain. A case is shown in Table~\ref{tab:case}.

\paragraph{DRPruning outperforms the baselines consistently across the domains.}
As shown in Table~\ref{tab:domain}, our method consistently outperforms the Constant and Sheared Llama scheduling strategies, achieving better downstream performance in most domains. This demonstrates that our approach not only improves learning in hard domains but also preserves or enhances performance in others. Additionally, our benchmark results align well with the average performance across 15 tasks, performing slightly below the open-source Sheared Llama 2.7B model. This consistency confirms the quality and reliability of our benchmark and results. Furthermore, we demonstrate that DRPruning benefits from finer domain segmentatio in Appendix~\ref{apx:finegrained}.

\begin{figure*}[!t]
 \centering
 \begin{minipage}{0.33\textwidth}
  \centering
  \includegraphics[width=\linewidth]{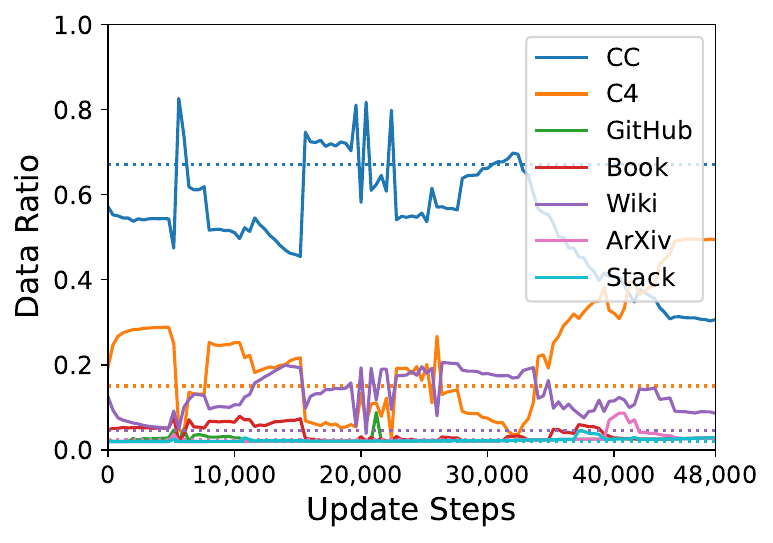}
 \end{minipage}\hfill
 \begin{minipage}{0.33\textwidth}
  \centering
  \includegraphics[width=\linewidth]{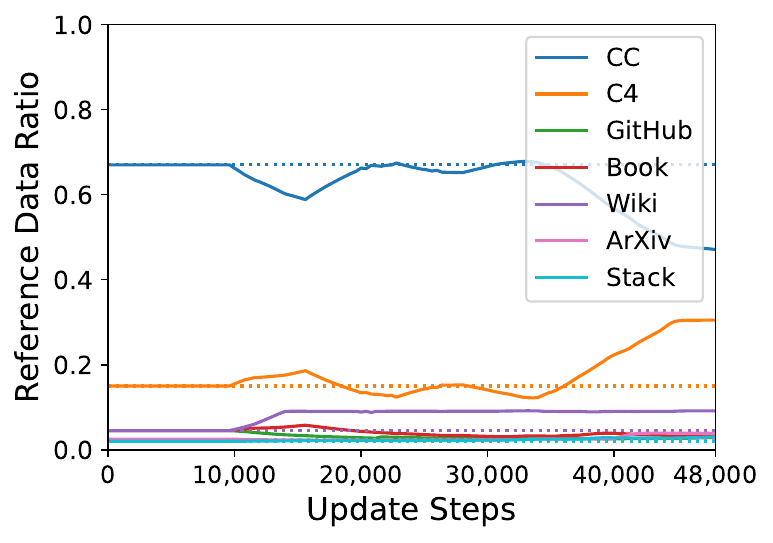}
 \end{minipage}\hfill
 \begin{minipage}{0.33\textwidth}
  \centering
  \includegraphics[width=\linewidth]{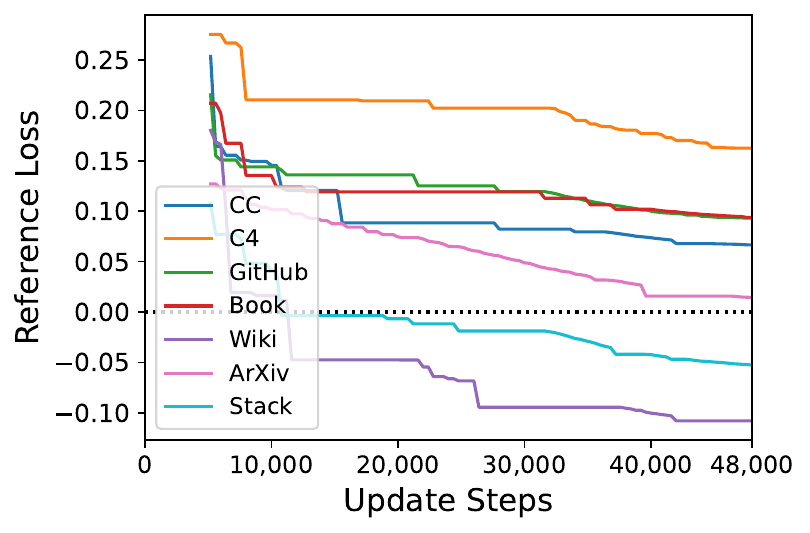}
 \end{minipage}
 \begin{minipage}{0.33\textwidth}
  \centering
  \includegraphics[width=\linewidth]{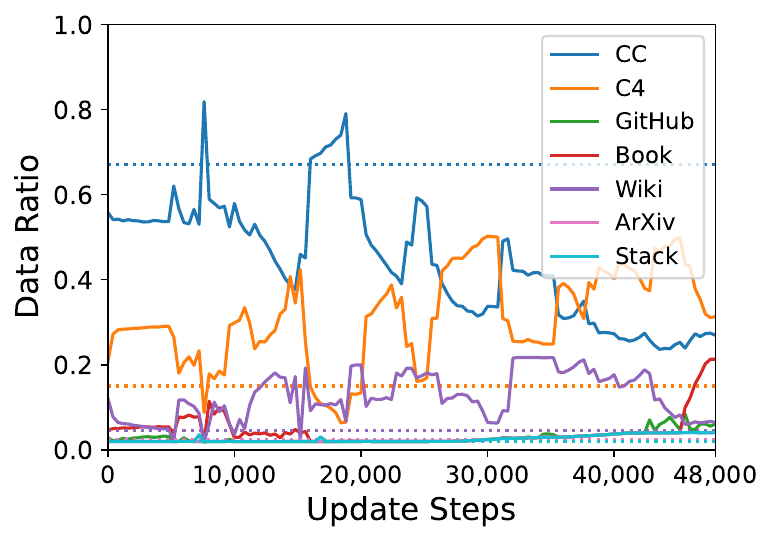}
 \end{minipage}\hfill
 \begin{minipage}{0.33\textwidth}
  \centering
  \includegraphics[width=\linewidth]{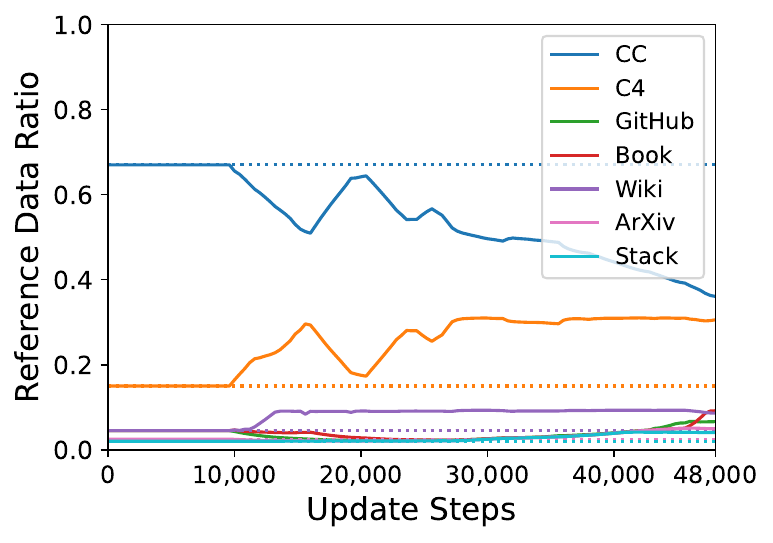}
 \end{minipage}\hfill
 \begin{minipage}{0.33\textwidth}
  \centering
  \includegraphics[width=\linewidth]{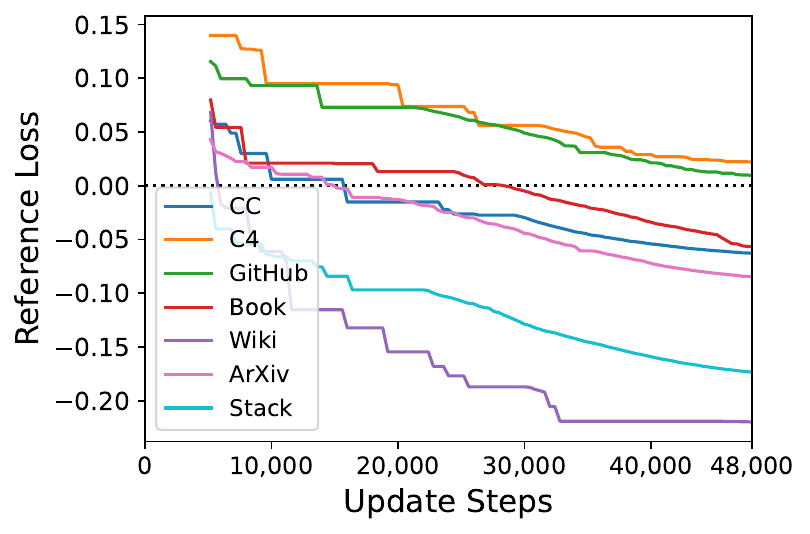}
 \end{minipage}
 \vspace{-6pt}
 \caption{\label{fig:hyperparameter}
 Variation of data ratio, reference data ratio, and reference loss during continued pretraining. The top three plots show results for pruning to 1.3B parameters, while the bottom three are for pruning to 2.7B parameters. Dashed lines in data ratio plots represent the initial reference data ratio for each domain. Reference loss plots display the difference from the initial value, with the dashed line at $y=0$ indicating no change from the initial ones.}
 \vspace{-6pt}
\end{figure*}

\subsection{Robustness under Distribution Shifts}

To verify our method under larger distribution shifts, we conduct experiments under the multilingual setting. To ensure a fair comparison, we reproduce DRPruning and ReSheared methods under the same experimental setup, detailed as follows:

\paragraph{Experimental setups.} We use Qwen2 series models \citep{Qwen2Technical_YYH+24}, which demonstrate superior multilingual performance, as the base models and explore two approaches: (1) continued pretraining from Qwen2-1.5B, and (2) pruning Qwen2-7B and then continued pretraining. Due to grouped query attention differences, we keep the head dimension unchanged, resulting in a 1.8B target architecture. We use the CulturaX dataset \citep{CulturaXCleaned_NVL+23} and select eight languages covered by Qwen2. The reference loss is initialized with Qwen2-7B on the validation set. For the reference ratio, we follow \citet{UnsupervisedCrosslingual_CKG+20}, upsampling low-resource languages with a smoothing rate of 0.3. We select various downstream tasks and report average performance. Detailed model configurations and metrics are provided in Appendix~\ref{apx:settingMultilingual}.

\paragraph{Our method demonstrates superior distribution robustness compared to ReSheared.}
As shown in Table~\ref{tab:multi}, our method outperforms ReSheared with an average gain of +1.50, and Qwen2-1.5B with +2.98. To analyze the source of these gains, as shown in Table~\ref{tab:multi_analysis}, Sheared Llama focuses on hard-to-improve areas like English, where its loss remains above the reference loss, leading to a highly imbalanced data distribution. In contrast, our method dynamically identifies underperforming domains, increasing the reference loss for high-resource languages and lowering it for low-resource ones. This leads to more balanced data scheduling and better evaluation loss, further demonstrating the robustness of our approach in handling distribution shifts.

\paragraph{Continued pretraining from the pruned model underperforms from the pretrained ones.}
Continued pretraining from the pruned model result in a slight performance drop (average -0.48), despite the increased parameters (1.8B vs. 1.5B). This contrasts with Sheared Llama, where continued pretraining on a smaller model shows minimal gains. In our case, using different data and a stronger small model improves performance during continued pretraining, leading to different results. Furthermore, DRPruning demonstrates efficacy not only under continued pretraining but also under instruction tuning, detailed in Appendix~\ref{apx:instuning}.

\subsection{Analysis of Hyperparameter Adjustment}

We analyze how our strategy adjusts training parameters, including the data ratio, reference data ratio, and reference loss, in the main experiment. The results are shown in Figure~\ref{fig:hyperparameter}.

\paragraph{High reliability of our approach.}
The trends for the 1.3B and 2.7B targets are similar, with increased allocation to C4 and Wiki and reduced allocation to CC. This highlights limitations in the current hyperparameter settings while confirming the reliability of our dynamic scheduling. In the later training stages, the CC domain exhibits lower potential with slower loss convergence, prompting our strategy to reduce its weight and increase the weight for C4. Wiki data consistently shows higher potential, leading to a significantly higher reference data ratio and the largest reference loss reduction.

\paragraph{Effective real-time evaluation of reference loss.}
To accelerate convergence, we select the minimum predicted value, which raises concerns about the inability to increase the reference loss when domain potential decreases. However, the two rightmost figures show that our predictions are conservative and decrease gradually during training. When potential declines, the rate of decrease slows, resulting in a relatively higher reference loss. This favorable outcome arises because we use loss from a limited training duration instead of the fully trained loss used in scaling laws, leading to more cautious estimates. Further, the similar trends between the 1.3B and 2.7B models indicate that our method provides reasonable training expectations, potentially supporting model training across broader ranges.

\section{Related Work}

\paragraph{LLM Pruning.}
Unstructured pruning \citep{LotteryTicket_FC19, SparseGPTMassive_FA23, SimpleEffective_SLBK23} removes individual weights but offers limited speedup. This study focuses on structured pruning \citep{DeepCompression_HMD16, LearningStructured_WWW+16}, which removes entire structural components, making it more effective for improving efficiency~\cite{liu-etal-2024-curriculum, pesf-kd}. 
In task-specific models, extensive pruning can retain performance \citep{LearningEfficient_LLS+17, PickingWinning_WZG20, BlockPruning_LCSR21, StructuredPruning_XZC22, ZipLMInferenceAware_KFA23}. However, for LLMs, as training data increases \citep{TrainingComputeOptimal_HBM+22}, fewer redundant parameters remain, leading to significant performance degradation after pruning.
To counter this, performance recovery techniques like continued pretraining are essential \citep{LLMPrunerStructural_MFW23, LoRAPruneStructured_ZCS+24}. However, continued pretraining of pruned models reduces loss at different rates across domains, resulting in less efficient data utilization \citep{ShearedLLaMA_XGZC23}. To address this, DRPruning dynamically adjusts the data distribution during training, ensuring balanced performance across domains.

\paragraph{Distributionally robust optimization (DRO).}
Overparameterized neural networks excel on i.i.d. test sets but struggle with underrepresented data groups \citep{TaggingPerformance_HS15, DemographicDialectal_BGO16, GenderDialect_Tat17}. Unlike empirical risk minimization, which minimizes expected loss for a fixed distribution, MultiDDS \citep{BalancingTraining_WTN20} optimizes the sampling distribution via gradient-based meta-learning but incurs higher computational and memory costs. In contrast, DRO \citep{DistributionallyRobust_DY10, RobustSolutions_BHW+12, DataDrivenRobust_BGK14} improves performance without additional complexity \citep{FairnessDemographics_HSNL18}.

DRO finds a model that performs well across multiple possible test distributions. Group DRO \citep{DistributionallyRobust_SKHL20} minimizes the worst-case loss over all domains without constraining potential distribution, while CVaR-Group DRO \citep{DistributionallyRobust_OSHL19} averages the largest $N$ group losses. These methods can be overly conservative, as they account for robustness to arbitrary subpopulations. \citet{DistributionallyRobust_ZLL+21} address this by constraining potential distribution within an $f$-divergence ball \citep{Csiszar67} around a  reference data ratio, yielding promising results \citep{TencentsMultilingual_JTL+22}.

\paragraph{DRO enhancement.}
DRO shows strong performance but relies on two main hyperparameters. The first is the reference loss, usually set by training an additional baseline model \citep{DistributionallyRobust_ZLL+21, TencentsMultilingual_JTL+22}, though this is expensive for LLMs. Sheared Llama uses scaling laws of model size to predict the pruned model's performance.
The second hyperparameter is the reference data ratio, often determined through temperature-based sampling \citep{MassivelyMultilingual_ABF+19, UnsupervisedCrosslingual_CKG+20} or manually \citep{LlamaOpen_TLI+23, Nemotron415B_PPJ+24}. However, fixed ratios can hinder model convergence in challenging distributions.
DRPruning shifts weight toward higher-loss domains, enhancing distribution robustness and improving downstream performance.

\section{Conclusion}
This paper presents DRPruning, a distributionally robust pruning method that addresses uneven performance degradation across domains during structured pruning. By utilizing and further improving distributionally robust optimization (DRO), our pruning method focuses more on domains with poorer performance, significantly accelerating performance recovery. 
It outperforms existing models and data scheduling methods in both monolingual and multilingual settings, achieving lower perplexity, higher task accuracy, and better instruction tuning outcomes. 
Further analysis demonstrates the robustness of our method against various domains and distribution shifts. Additionally, the dynamic adjustment of reference loss and data ratios exhibits broad applicability, with strong potential to support balanced training across diverse tasks.

\section*{Limitations} 

\paragraph{Exploration of smaller pruning ratios.} Due to computational constraints, we are unable to explore pruning to larger models, i.e., employing smaller pruning ratios. Retaining a larger proportion of the model's parameters may lead to different outcomes in some experiments. For example, it remains to be investigated whether pruning larger models provides benefits, and whether it is better to continue pretraining from a pruned model or from a smaller, fully pretrained model.

\paragraph{More extensive continued pretraining.}
\citet{ShearedLLaMA_XGZC23} point out that pruned models exhibit higher training ceilings. Although good performance can be achieved with tens of billions of training samples, this study does not investigate whether training the models to full convergence using hundreds or thousands of billions of samples would yield better results than continuing pretraining from existing pretrained models under similar settings.

\paragraph{Validation in other scenarios.} We have validated our method's effectiveness in the pruning phase, the pruning recovery phase, the continued pretraining phase, and the instruction tuning phase in Appendix~\ref{apx:instuning}. Our method is expected to be applicable in broader contexts, such as pretraining from scratch and cross-domain post-training. Broader validation would further demonstrate the superiority of our approach.

\section*{Ethics Statement}
Our work adheres to the ACL Ethics Policy and uses publicly available datasets for reproducibility. LLMs may exhibit racial and gender biases, so we strongly recommend users assess potential biases before applying the models in specific contexts. Additionally, due to the difficulty of controlling LLM outputs, users should be cautious of issues arising from hallucinations.

\section*{Acknowledgments}
This work was supported in part by the National Natural Science Foundation of China (Grant No. 62206076), Guangdong S\&T Program (Grant No. 2024B0101050003), Guangdong Basic and Applied Basic Research Foundation (Grant No. 2024A1515011491), and Shenzhen Science and Technology Program (Grant Nos. ZDSYS20230626091203008, KJZD20231023094700001, KQTD2024072910215406). We would like to thank the anonymous reviewers and meta-reviewer for their insightful suggestions.

\bibliography{custom}

\begin{thebibliography}{80}
\expandafter\ifx\csname natexlab\endcsname\relax\def\natexlab#1{#1}\fi

\bibitem[{Arivazhagan et~al.(2019)Arivazhagan, Bapna, Firat, Lepikhin, Johnson, Krikun, Chen, Cao, Foster, Cherry, Macherey, Chen, and Wu}]{MassivelyMultilingual_ABF+19}
Naveen Arivazhagan, Ankur Bapna, Orhan Firat, Dmitry Lepikhin, Melvin Johnson, Maxim Krikun, Mia~Xu Chen, Yuan Cao, George Foster, Colin Cherry, Wolfgang Macherey, Zhifeng Chen, and Yonghui Wu. 2019.
\newblock \href {https://arxiv.org/abs/1907.05019} {Massively {{Multilingual Neural Machine Translation}} in the {{Wild}}: {{Findings}} and {{Challenges}}}.

\bibitem[{{Ben-Tal} et~al.(2012){Ben-Tal}, den Hertog, Waegenaere, Melenberg, and Rennen}]{RobustSolutions_BHW+12}
Aharon {Ben-Tal}, Dick den Hertog, Anja~De Waegenaere, Bertrand Melenberg, and Gijs Rennen. 2012.
\newblock \href {https://doi.org/10.1287/mnsc.1120.1641} {Robust {{Solutions}} of {{Optimization Problems Affected}} by {{Uncertain Probabilities}}}.
\newblock \emph{Management Science}.

\bibitem[{Bertsimas et~al.(2014)Bertsimas, Gupta, and Kallus}]{DataDrivenRobust_BGK14}
Dimitris Bertsimas, Vishal Gupta, and Nathan Kallus. 2014.
\newblock \href {https://doi.org/10.48550/arXiv.1401.0212} {Data-{{Driven Robust Optimization}}}.

\bibitem[{Bian et~al.(2024)Bian, Han, Sun, Lin, Lu, He, Jiang, and Dong}]{ChatGPTKnowledgeableInexperienced_2023}
Ning Bian, Xianpei Han, Le~Sun, Hongyu Lin, Yaojie Lu, Ben He, Shanshan Jiang, and Bin Dong. 2024.
\newblock \href {https://aclanthology.org/2024.lrec-main.276} {Chatgpt is a knowledgeable but inexperienced solver: An investigation of commonsense problem in large language models}.
\newblock In \emph{Proceedings of the 2024 Joint International Conference on Computational Linguistics, Language Resources and Evaluation, {LREC/COLING} 2024, 20-25 May, 2024, Torino, Italy}, pages 3098--3110. {ELRA} and {ICCL}.

\bibitem[{Biderman et~al.(2023)Biderman, Schoelkopf, Anthony, Bradley, O'Brien, Hallahan, Khan, Purohit, Prashanth, Raff, Skowron, Sutawika, and van~der Wal}]{PythiaSuite_BSA+23}
Stella Biderman, Hailey Schoelkopf, Quentin~Gregory Anthony, Herbie Bradley, Kyle O'Brien, Eric Hallahan, Mohammad~Aflah Khan, Shivanshu Purohit, USVSN~Sai Prashanth, Edward Raff, Aviya Skowron, Lintang Sutawika, and Oskar van~der Wal. 2023.
\newblock \href {https://proceedings.mlr.press/v202/biderman23a.html} {Pythia: {A} suite for analyzing large language models across training and scaling}.
\newblock In \emph{International Conference on Machine Learning, {ICML} 2023, 23-29 July 2023, Honolulu, Hawaii, {USA}}, volume 202 of \emph{Proceedings of Machine Learning Research}, pages 2397--2430. {PMLR}.

\bibitem[{Bisk et~al.(2020)Bisk, Zellers, Bras, Gao, and Choi}]{PIQAReasoning_BZB+19}
Yonatan Bisk, Rowan Zellers, Ronan~Le Bras, Jianfeng Gao, and Yejin Choi. 2020.
\newblock \href {https://doi.org/10.1609/AAAI.V34I05.6239} {{PIQA:} reasoning about physical commonsense in natural language}.
\newblock In \emph{The Thirty-Fourth {AAAI} Conference on Artificial Intelligence, {AAAI} 2020, The Thirty-Second Innovative Applications of Artificial Intelligence Conference, {IAAI} 2020, The Tenth {AAAI} Symposium on Educational Advances in Artificial Intelligence, {EAAI} 2020, New York, NY, USA, February 7-12, 2020}, pages 7432--7439. {AAAI} Press.

\bibitem[{Blodgett et~al.(2016)Blodgett, Green, and O{'}Connor}]{DemographicDialectal_BGO16}
Su~Lin Blodgett, Lisa Green, and Brendan O{'}Connor. 2016.
\newblock \href {https://doi.org/10.18653/v1/D16-1120} {Demographic dialectal variation in social media: A case study of {A}frican-{A}merican {E}nglish}.
\newblock In \emph{Proceedings of the 2016 Conference on Empirical Methods in Natural Language Processing}, pages 1119--1130, Austin, Texas. Association for Computational Linguistics.

\bibitem[{Clark et~al.(2019)Clark, Lee, Chang, Kwiatkowski, Collins, and Toutanova}]{BoolQExploring_CLC+19}
Christopher Clark, Kenton Lee, Ming-Wei Chang, Tom Kwiatkowski, Michael Collins, and Kristina Toutanova. 2019.
\newblock \href {https://doi.org/10.18653/v1/N19-1300} {{B}ool{Q}: Exploring the surprising difficulty of natural yes/no questions}.
\newblock In \emph{Proceedings of the 2019 Conference of the North {A}merican Chapter of the Association for Computational Linguistics: Human Language Technologies, Volume 1 (Long and Short Papers)}, pages 2924--2936, Minneapolis, Minnesota. Association for Computational Linguistics.

\bibitem[{Clark et~al.(2018)Clark, Cowhey, Etzioni, Khot, Sabharwal, Schoenick, and Tafjord}]{ThinkYou_CCE+18}
Peter Clark, Isaac Cowhey, Oren Etzioni, Tushar Khot, Ashish Sabharwal, Carissa Schoenick, and Oyvind Tafjord. 2018.
\newblock \href {https://arxiv.org/abs/1803.05457} {Think you have {{Solved Question Answering}}? {{Try ARC}}, the {{AI2 Reasoning Challenge}}}.

\bibitem[{Computer(2023)}]{together2023redpajama}
Together Computer. 2023.
\newblock \href {https://github.com/togethercomputer/RedPajama-Data} {Redpajama: an open dataset for training large language models}.

\bibitem[{Conneau et~al.(2020)Conneau, Khandelwal, Goyal, Chaudhary, Wenzek, Guzm{\'a}n, Grave, Ott, Zettlemoyer, and Stoyanov}]{UnsupervisedCrosslingual_CKG+20}
Alexis Conneau, Kartikay Khandelwal, Naman Goyal, Vishrav Chaudhary, Guillaume Wenzek, Francisco Guzm{\'a}n, Edouard Grave, Myle Ott, Luke Zettlemoyer, and Veselin Stoyanov. 2020.
\newblock \href {https://doi.org/10.18653/v1/2020.acl-main.747} {Unsupervised cross-lingual representation learning at scale}.
\newblock In \emph{Proceedings of the 58th Annual Meeting of the Association for Computational Linguistics}, pages 8440--8451, Online. Association for Computational Linguistics.

\bibitem[{Conneau et~al.(2018)Conneau, Rinott, Lample, Williams, Bowman, Schwenk, and Stoyanov}]{XNLIEvaluating_CRL+18}
Alexis Conneau, Ruty Rinott, Guillaume Lample, Adina Williams, Samuel Bowman, Holger Schwenk, and Veselin Stoyanov. 2018.
\newblock \href {https://doi.org/10.18653/v1/D18-1269} {{XNLI}: Evaluating cross-lingual sentence representations}.
\newblock In \emph{Proceedings of the 2018 Conference on Empirical Methods in Natural Language Processing}, pages 2475--2485, Brussels, Belgium. Association for Computational Linguistics.

\bibitem[{Csisz{\'a}r(1967)}]{Csiszar67}
Imre Csisz{\'a}r. 1967.
\newblock \href {https://www.scirp.org/reference/referencespapers?referenceid=1866686} {Information-type measures of difference of probability distributions and indirect observation}.
\newblock \emph{Studia Scientifica Mathematica Hungary}, 2:299--318.

\bibitem[{Dao et~al.(2022)Dao, Fu, Ermon, Rudra, and R{\'{e}}}]{FlashAttentionFast_DFE+22}
Tri Dao, Daniel~Y. Fu, Stefano Ermon, Atri Rudra, and Christopher R{\'{e}}. 2022.
\newblock \href {http://papers.nips.cc/paper\_files/paper/2022/hash/67d57c32e20fd0a7a302cb81d36e40d5-Abstract-Conference.html} {Flashattention: Fast and memory-efficient exact attention with io-awareness}.
\newblock In \emph{Advances in Neural Information Processing Systems 35: Annual Conference on Neural Information Processing Systems 2022, NeurIPS 2022, New Orleans, LA, USA, November 28 - December 9, 2022}.

\bibitem[{DeepSeek-AI(2025)}]{deepseekai2025deepseekr1incentivizingreasoningcapability}
DeepSeek-AI. 2025.
\newblock \href {http://arxiv.org/abs/2501.12948} {Deepseek-r1: Incentivizing reasoning capability in llms via reinforcement learning}.

\bibitem[{Delage and Ye(2010)}]{DistributionallyRobust_DY10}
Erick Delage and Yinyu Ye. 2010.
\newblock \href {https://doi.org/10.1287/opre.1090.0741} {Distributionally {{Robust Optimization Under Moment Uncertainty}} with {{Application}} to {{Data-Driven Problems}}}.
\newblock \emph{Operations Research}, 58(3):595--612.

\bibitem[{Deng et~al.(2023)Deng, Ding, Liu, Zhang, Tao, and Zhang}]{ImprovingSimultaneous_DDL+23}
Hexuan Deng, Liang Ding, Xuebo Liu, Meishan Zhang, Dacheng Tao, and Min Zhang. 2023.
\newblock \href {https://doi.org/10.1609/AAAI.V37I11.26497} {Improving simultaneous machine translation with monolingual data}.
\newblock In \emph{Thirty-Seventh {AAAI} Conference on Artificial Intelligence, {AAAI} 2023, Thirty-Fifth Conference on Innovative Applications of Artificial Intelligence, {IAAI} 2023, Thirteenth Symposium on Educational Advances in Artificial Intelligence, {EAAI} 2023, Washington, DC, USA, February 7-14, 2023}, pages 12728--12736. {AAAI} Press.

\bibitem[{Deng et~al.(2024)Deng, Jiao, Liu, Zhang, and Tu}]{NewTermBenchmarking_DJL+24}
Hexuan Deng, Wenxiang Jiao, Xuebo Liu, Min Zhang, and Zhaopeng Tu. 2024.
\newblock \href {https://proceedings.neurips.cc/paper_files/paper/2024/file/3eec719ab86712d32b065c5977f94ad0-Paper-Datasets_and_Benchmarks_Track.pdf} {Newterm: Benchmarking real-time new terms for large language models with annual updates}.
\newblock In \emph{Advances in Neural Information Processing Systems}, volume~37, pages 35760--35795. Curran Associates, Inc.

\bibitem[{Frankle and Carbin(2019)}]{LotteryTicket_FC19}
Jonathan Frankle and Michael Carbin. 2019.
\newblock \href {https://openreview.net/forum?id=rJl-b3RcF7} {The lottery ticket hypothesis: Finding sparse, trainable neural networks}.
\newblock In \emph{7th International Conference on Learning Representations, {ICLR} 2019, New Orleans, LA, USA, May 6-9, 2019}. OpenReview.net.

\bibitem[{Frantar and Alistarh(2023)}]{SparseGPTMassive_FA23}
Elias Frantar and Dan Alistarh. 2023.
\newblock \href {https://proceedings.mlr.press/v202/frantar23a.html} {Sparsegpt: Massive language models can be accurately pruned in one-shot}.
\newblock In \emph{International Conference on Machine Learning, {ICML} 2023, 23-29 July 2023, Honolulu, Hawaii, {USA}}, volume 202 of \emph{Proceedings of Machine Learning Research}, pages 10323--10337. {PMLR}.

\bibitem[{Frieder et~al.(2023)Frieder, Pinchetti, Chevalier, Griffiths, Salvatori, Lukasiewicz, Petersen, and Berner}]{MathematicalCapabilitiesChatGPT_2023}
Simon Frieder, Luca Pinchetti, Alexis Chevalier, Ryan{-}Rhys Griffiths, Tommaso Salvatori, Thomas Lukasiewicz, Philipp Petersen, and Julius Berner. 2023.
\newblock \href {http://papers.nips.cc/paper\_files/paper/2023/hash/58168e8a92994655d6da3939e7cc0918-Abstract-Datasets\_and\_Benchmarks.html} {Mathematical capabilities of chatgpt}.
\newblock In \emph{Advances in Neural Information Processing Systems 36: Annual Conference on Neural Information Processing Systems 2023, NeurIPS 2023, New Orleans, LA, USA, December 10 - 16, 2023}.

\bibitem[{Fudenberg and Levine(1998)}]{fudenberg1998theory}
Drew Fudenberg and David~K Levine. 1998.
\newblock \href {https://books.google.com.tw/books?hl=zh-CN&lr=&id=G6vTQFluxuEC&oi=fnd&pg=PR11&dq=The+theory+of+learning+in+games&ots=S_FBMM6w7-&sig=lOuW2WPzha3Mla8ajCPFBVYGQWs&redir_esc=y#v=onepage&q=The%20theory%20of%20learning%20in%20games&f=false} {\emph{The theory of learning in games}}, volume~2.
\newblock MIT press.

\bibitem[{Gao et~al.(2024)Gao, Tow, Abbasi, Biderman, Black, DiPofi, Foster, Golding, Hsu, Le~Noac'h, Li, McDonell, Muennighoff, Ociepa, Phang, Reynolds, Schoelkopf, Skowron, Sutawika, Tang, Thite, Wang, Wang, and Zou}]{eval-harness}
Leo Gao, Jonathan Tow, Baber Abbasi, Stella Biderman, Sid Black, Anthony DiPofi, Charles Foster, Laurence Golding, Jeffrey Hsu, Alain Le~Noac'h, Haonan Li, Kyle McDonell, Niklas Muennighoff, Chris Ociepa, Jason Phang, Laria Reynolds, Hailey Schoelkopf, Aviya Skowron, Lintang Sutawika, Eric Tang, Anish Thite, Ben Wang, Kevin Wang, and Andy Zou. 2024.
\newblock \href {https://doi.org/10.5281/zenodo.12608602} {A framework for few-shot language model evaluation}.

\bibitem[{Ghorbani et~al.(2022)Ghorbani, Firat, Freitag, Bapna, Krikun, Garcia, Chelba, and Cherry}]{ScalingLaws_GFF+21}
Behrooz Ghorbani, Orhan Firat, Markus Freitag, Ankur Bapna, Maxim Krikun, Xavier Garcia, Ciprian Chelba, and Colin Cherry. 2022.
\newblock \href {https://openreview.net/forum?id=hR\_SMu8cxCV} {Scaling laws for neural machine translation}.
\newblock In \emph{The Tenth International Conference on Learning Representations, {ICLR} 2022, Virtual Event, April 25-29, 2022}. OpenReview.net.

\bibitem[{Goyal et~al.(2022)Goyal, Gao, Chaudhary, Chen, Wenzek, Ju, Krishnan, Ranzato, Guzm{\'{a}}n, and Fan}]{nllb2022}
Naman Goyal, Cynthia Gao, Vishrav Chaudhary, Peng{-}Jen Chen, Guillaume Wenzek, Da~Ju, Sanjana Krishnan, Marc'Aurelio Ranzato, Francisco Guzm{\'{a}}n, and Angela Fan. 2022.
\newblock \href {https://doi.org/10.1162/TACL\_A\_00474} {The flores-101 evaluation benchmark for low-resource and multilingual machine translation}.
\newblock \emph{Trans. Assoc. Comput. Linguistics}, 10:522--538.

\bibitem[{Han et~al.(2015)Han, Mao, and Dally}]{DeepCompression_HMD16}
Song Han, Huizi Mao, and William~J. Dally. 2015.
\newblock \href {https://arxiv.org/abs/1510.00149} {Deep {{Compression}}: {{Compressing Deep Neural Networks}} with {{Pruning}}, {{Trained Quantization}} and {{Huffman Coding}}}.

\bibitem[{Hashimoto et~al.(2018)Hashimoto, Srivastava, Namkoong, and Liang}]{FairnessDemographics_HSNL18}
Tatsunori~B. Hashimoto, Megha Srivastava, Hongseok Namkoong, and Percy Liang. 2018.
\newblock \href {http://proceedings.mlr.press/v80/hashimoto18a.html} {Fairness without demographics in repeated loss minimization}.
\newblock In \emph{Proceedings of the 35th International Conference on Machine Learning, {ICML} 2018, Stockholmsm{\"{a}}ssan, Stockholm, Sweden, July 10-15, 2018}, volume~80 of \emph{Proceedings of Machine Learning Research}, pages 1934--1943. {PMLR}.

\bibitem[{Hendrycks et~al.(2021)Hendrycks, Burns, Basart, Zou, Mazeika, Song, and Steinhardt}]{MeasuringMassive_HBB+21}
Dan Hendrycks, Collin Burns, Steven Basart, Andy Zou, Mantas Mazeika, Dawn Song, and Jacob Steinhardt. 2021.
\newblock \href {https://openreview.net/forum?id=d7KBjmI3GmQ} {Measuring massive multitask language understanding}.
\newblock In \emph{9th International Conference on Learning Representations, {ICLR} 2021, Virtual Event, Austria, May 3-7, 2021}. OpenReview.net.

\bibitem[{Hoffmann et~al.(2022{\natexlab{a}})Hoffmann, Borgeaud, Mensch, Buchatskaya, Cai, Rutherford, Casas, Hendricks, Welbl, Clark, Hennigan, Noland, Millican, van~den Driessche, Damoc, Guy, Osindero, Simonyan, Elsen, Rae, Vinyals, and Sifre}]{TrainingComputeOptimal_HBM+22a}
Jordan Hoffmann, Sebastian Borgeaud, Arthur Mensch, Elena Buchatskaya, Trevor Cai, Eliza Rutherford, Diego de~Las Casas, Lisa~Anne Hendricks, Johannes Welbl, Aidan Clark, Tom Hennigan, Eric Noland, Katie Millican, George van~den Driessche, Bogdan Damoc, Aurelia Guy, Simon Osindero, Karen Simonyan, Erich Elsen, Jack~W. Rae, Oriol Vinyals, and Laurent Sifre. 2022{\natexlab{a}}.
\newblock \href {https://arxiv.org/abs/2203.15556} {Training {{Compute-Optimal Large Language Models}}}.

\bibitem[{Hoffmann et~al.(2022{\natexlab{b}})Hoffmann, Borgeaud, Mensch, Buchatskaya, Cai, Rutherford, Casas, Hendricks, Welbl, Clark, Hennigan, Noland, Millican, van~den Driessche, Damoc, Guy, Osindero, Simonyan, Elsen, Rae, Vinyals, and Sifre}]{TrainingComputeOptimal_HBM+22}
Jordan Hoffmann, Sebastian Borgeaud, Arthur Mensch, Elena Buchatskaya, Trevor Cai, Eliza Rutherford, Diego de~Las Casas, Lisa~Anne Hendricks, Johannes Welbl, Aidan Clark, Tom Hennigan, Eric Noland, Katie Millican, George van~den Driessche, Bogdan Damoc, Aurelia Guy, Simon Osindero, Karen Simonyan, Erich Elsen, Jack~W. Rae, Oriol Vinyals, and Laurent Sifre. 2022{\natexlab{b}}.
\newblock \href {https://arxiv.org/abs/2203.15556} {Training {{Compute-Optimal Large Language Models}}}.

\bibitem[{Hovy and S{\o}gaard(2015)}]{TaggingPerformance_HS15}
Dirk Hovy and Anders S{\o}gaard. 2015.
\newblock \href {https://doi.org/10.3115/v1/P15-2079} {Tagging performance correlates with author age}.
\newblock In \emph{Proceedings of the 53rd Annual Meeting of the Association for Computational Linguistics and the 7th International Joint Conference on Natural Language Processing (Volume 2: Short Papers)}, pages 483--488, Beijing, China. Association for Computational Linguistics.

\bibitem[{Jiao et~al.(2023{\natexlab{a}})Jiao, Huang, Wang, He, Liang, Wang, Shi, and Tu}]{ParroTTranslating_JHW+23}
Wenxiang Jiao, Jen-tse Huang, Wenxuan Wang, Zhiwei He, Tian Liang, Xing Wang, Shuming Shi, and Zhaopeng Tu. 2023{\natexlab{a}}.
\newblock \href {https://doi.org/10.18653/v1/2023.findings-emnlp.1001} {{P}arro{T}: Translating during chat using large language models tuned with human translation and feedback}.
\newblock In \emph{Findings of the Association for Computational Linguistics: EMNLP 2023}, pages 15009--15020, Singapore. Association for Computational Linguistics.

\bibitem[{Jiao et~al.(2022)Jiao, Tu, Li, Wang, Huang, and Shi}]{TencentsMultilingual_JTL+22}
Wenxiang Jiao, Zhaopeng Tu, Jiarui Li, Wenxuan Wang, Jen-tse Huang, and Shuming Shi. 2022.
\newblock \href {https://aclanthology.org/2022.wmt-1.102} {Tencent{'}s multilingual machine translation system for {WMT}22 large-scale {A}frican languages}.
\newblock In \emph{Proceedings of the Seventh Conference on Machine Translation (WMT)}, pages 1049--1056, Abu Dhabi, United Arab Emirates (Hybrid). Association for Computational Linguistics.

\bibitem[{Jiao et~al.(2023{\natexlab{b}})Jiao, Wang, Huang, Wang, Shi, and Tu}]{ChatGPTGoodTranslator_2023}
Wenxiang Jiao, Wenxuan Wang, Jen-tse Huang, Xing Wang, Shuming Shi, and Zhaopeng Tu. 2023{\natexlab{b}}.
\newblock \href {https://arxiv.org/abs/2301.08745} {Is {{ChatGPT A Good Translator}}? {{A Preliminary Study}}}.
\newblock \emph{ArXiv preprint}, abs/2301.08745.

\bibitem[{Joshi et~al.(2017)Joshi, Choi, Weld, and Zettlemoyer}]{TriviaQALarge_JCWZ17}
Mandar Joshi, Eunsol Choi, Daniel Weld, and Luke Zettlemoyer. 2017.
\newblock \href {https://doi.org/10.18653/v1/P17-1147} {{T}rivia{QA}: A large scale distantly supervised challenge dataset for reading comprehension}.
\newblock In \emph{Proceedings of the 55th Annual Meeting of the Association for Computational Linguistics (Volume 1: Long Papers)}, pages 1601--1611, Vancouver, Canada. Association for Computational Linguistics.

\bibitem[{Kaplan et~al.(2020)Kaplan, McCandlish, Henighan, Brown, Chess, Child, Gray, Radford, Wu, and Amodei}]{ScalingLaws_KMH+20}
Jared Kaplan, Sam McCandlish, Tom Henighan, Tom~B. Brown, Benjamin Chess, Rewon Child, Scott Gray, Alec Radford, Jeffrey Wu, and Dario Amodei. 2020.
\newblock \href {https://arxiv.org/abs/2001.08361} {Scaling {{Laws}} for {{Neural Language Models}}}.
\newblock \emph{ArXiv preprint}, abs/2001.08361.

\bibitem[{Kocijan et~al.(2020)Kocijan, Lukasiewicz, Davis, Marcus, and Morgenstern}]{ReviewWinograd_KLD+20}
Vid Kocijan, Thomas Lukasiewicz, Ernest Davis, Gary Marcus, and Leora Morgenstern. 2020.
\newblock \href {https://arxiv.org/abs/2004.13831} {A {{Review}} of {{Winograd Schema Challenge Datasets}} and {{Approaches}}}.

\bibitem[{Kocmi et~al.(2023)Kocmi, Avramidis, Bawden, Bojar, Dvorkovich, Federmann, Fishel, Freitag, Gowda, Grundkiewicz, Haddow, Koehn, Marie, Monz, Morishita, Murray, Nagata, Nakazawa, Popel, Popovic, and Shmatova}]{DBLP:conf/wmt/KocmiABBDFFFGGH23}
Tom Kocmi, Eleftherios Avramidis, Rachel Bawden, Ondrej Bojar, Anton Dvorkovich, Christian Federmann, Mark Fishel, Markus Freitag, Thamme Gowda, Roman Grundkiewicz, Barry Haddow, Philipp Koehn, Benjamin Marie, Christof Monz, Makoto Morishita, Kenton Murray, Makoto Nagata, Toshiaki Nakazawa, Martin Popel, Maja Popovic, and Mariya Shmatova. 2023.
\newblock \href {https://doi.org/10.18653/V1/2023.WMT-1.1} {Findings of the 2023 conference on machine translation {(WMT23):} llms are here but not quite there yet}.
\newblock In \emph{Proceedings of the Eighth Conference on Machine Translation, {WMT} 2023, Singapore, December 6-7, 2023}, pages 1--42. Association for Computational Linguistics.

\bibitem[{Kurtic et~al.(2023)Kurtic, Frantar, and Alistarh}]{ZipLMInferenceAware_KFA23}
Eldar Kurtic, Elias Frantar, and Dan Alistarh. 2023.
\newblock \href {http://papers.nips.cc/paper\_files/paper/2023/hash/ced46a50befedcb884ccf0cbe8c3ad23-Abstract-Conference.html} {Ziplm: Inference-aware structured pruning of language models}.
\newblock In \emph{Advances in Neural Information Processing Systems 36: Annual Conference on Neural Information Processing Systems 2023, NeurIPS 2023, New Orleans, LA, USA, December 10 - 16, 2023}.

\bibitem[{Kwiatkowski et~al.(2019)Kwiatkowski, Palomaki, Redfield, Collins, Parikh, Alberti, Epstein, Polosukhin, Devlin, Lee, Toutanova, Jones, Kelcey, Chang, Dai, Uszkoreit, Le, and Petrov}]{NaturalQuestions_KPR+19}
Tom Kwiatkowski, Jennimaria Palomaki, Olivia Redfield, Michael Collins, Ankur~P. Parikh, Chris Alberti, Danielle Epstein, Illia Polosukhin, Jacob Devlin, Kenton Lee, Kristina Toutanova, Llion Jones, Matthew Kelcey, Ming{-}Wei Chang, Andrew~M. Dai, Jakob Uszkoreit, Quoc Le, and Slav Petrov. 2019.
\newblock \href {https://doi.org/10.1162/TACL\_A\_00276} {Natural questions: a benchmark for question answering research}.
\newblock \emph{Trans. Assoc. Comput. Linguistics}, 7:452--466.

\bibitem[{Lagunas et~al.(2021)Lagunas, Charlaix, Sanh, and Rush}]{BlockPruning_LCSR21}
Fran{\c{c}}ois Lagunas, Ella Charlaix, Victor Sanh, and Alexander Rush. 2021.
\newblock \href {https://doi.org/10.18653/v1/2021.emnlp-main.829} {Block pruning for faster transformers}.
\newblock In \emph{Proceedings of the 2021 Conference on Empirical Methods in Natural Language Processing}, pages 10619--10629, Online and Punta Cana, Dominican Republic. Association for Computational Linguistics.

\bibitem[{Li et~al.(2024)Li, Chen, Han, and Bai}]{NutePruneEfficient_LCHB24}
Shengrui Li, Junzhe Chen, Xueting Han, and Jing Bai. 2024.
\newblock \href {https://arxiv.org/abs/2402.09773} {{{NutePrune}}: {{Efficient Progressive Pruning}} with {{Numerous Teachers}} for {{Large Language Models}}}.

\bibitem[{Lin et~al.(2022)Lin, Hilton, and Evans}]{TruthfulQAMeasuring_LHE22}
Stephanie Lin, Jacob Hilton, and Owain Evans. 2022.
\newblock \href {https://doi.org/10.18653/v1/2022.acl-long.229} {{T}ruthful{QA}: Measuring how models mimic human falsehoods}.
\newblock In \emph{Proceedings of the 60th Annual Meeting of the Association for Computational Linguistics (Volume 1: Long Papers)}, pages 3214--3252, Dublin, Ireland. Association for Computational Linguistics.

\bibitem[{Lin et~al.(2021)Lin, Mihaylov, Artetxe, Wang, Chen, Simig, Ott, Goyal, Bhosale, Du, Pasunuru, Shleifer, Koura, Chaudhary, O'Horo, Wang, Zettlemoyer, Kozareva, Diab, Stoyanov, and Li}]{FewshotLearning_LMA+22}
Xi~Victoria Lin, Todor Mihaylov, Mikel Artetxe, Tianlu Wang, Shuohui Chen, Daniel Simig, Myle Ott, Naman Goyal, Shruti Bhosale, Jingfei Du, Ramakanth Pasunuru, Sam Shleifer, Punit~Singh Koura, Vishrav Chaudhary, Brian O'Horo, Jeff Wang, Luke Zettlemoyer, Zornitsa Kozareva, Mona Diab, Veselin Stoyanov, and Xian Li. 2021.
\newblock \href {https://arxiv.org/abs/2112.10668} {Few-shot {{Learning}} with {{Multilingual Language Models}}}.

\bibitem[{Liu et~al.(2020)Liu, Cui, Liu, Huang, Wang, and Zhang}]{LogiQAChallenge_LCL+20}
Jian Liu, Leyang Cui, Hanmeng Liu, Dandan Huang, Yile Wang, and Yue Zhang. 2020.
\newblock \href {https://doi.org/10.24963/IJCAI.2020/501} {Logiqa: {A} challenge dataset for machine reading comprehension with logical reasoning}.
\newblock In \emph{Proceedings of the Twenty-Ninth International Joint Conference on Artificial Intelligence, {IJCAI} 2020}, pages 3622--3628. ijcai.org.

\bibitem[{Liu et~al.(2024)Liu, Liu, Lian, Cheng, Rao, Yu, Deng, and Zhang}]{liu-etal-2024-curriculum}
Liangxin Liu, Xuebo Liu, Lian Lian, Shengjun Cheng, Jun Rao, Tengfei Yu, Hexuan Deng, and Min Zhang. 2024.
\newblock \href {https://doi.org/10.18653/v1/2024.emnlp-main.768} {Curriculum consistency learning for conditional sentence generation}.
\newblock In \emph{Proceedings of the 2024 Conference on Empirical Methods in Natural Language Processing}, pages 13865--13881, Miami, Florida, USA. Association for Computational Linguistics.

\bibitem[{Liu et~al.(2017)Liu, Li, Shen, Huang, Yan, and Zhang}]{LearningEfficient_LLS+17}
Zhuang Liu, Jianguo Li, Zhiqiang Shen, Gao Huang, Shoumeng Yan, and Changshui Zhang. 2017.
\newblock \href {https://doi.org/10.1109/ICCV.2017.298} {Learning efficient convolutional networks through network slimming}.
\newblock In \emph{{IEEE} International Conference on Computer Vision, {ICCV} 2017, Venice, Italy, October 22-29, 2017}, pages 2755--2763. {IEEE} Computer Society.

\bibitem[{Louizos et~al.(2018)Louizos, Welling, and Kingma}]{LearningSparse_LWK18}
Christos Louizos, Max Welling, and Diederik~P. Kingma. 2018.
\newblock \href {https://openreview.net/forum?id=H1Y8hhg0b} {Learning sparse neural networks through l{\_}0 regularization}.
\newblock In \emph{6th International Conference on Learning Representations, {ICLR} 2018, Vancouver, BC, Canada, April 30 - May 3, 2018, Conference Track Proceedings}. OpenReview.net.

\bibitem[{Ma et~al.(2023)Ma, Fang, and Wang}]{LLMPrunerStructural_MFW23}
Xinyin Ma, Gongfan Fang, and Xinchao Wang. 2023.
\newblock \href {http://papers.nips.cc/paper\_files/paper/2023/hash/44956951349095f74492a5471128a7e0-Abstract-Conference.html} {Llm-pruner: On the structural pruning of large language models}.
\newblock In \emph{Advances in Neural Information Processing Systems 36: Annual Conference on Neural Information Processing Systems 2023, NeurIPS 2023, New Orleans, LA, USA, December 10 - 16, 2023}.

\bibitem[{Nguyen et~al.(2024)Nguyen, Nguyen, Lai, Man, Ngo, Dernoncourt, Rossi, and Nguyen}]{CulturaXCleaned_NVL+23}
Thuat Nguyen, Chien~Van Nguyen, Viet~Dac Lai, Hieu Man, Nghia~Trung Ngo, Franck Dernoncourt, Ryan~A. Rossi, and Thien~Huu Nguyen. 2024.
\newblock \href {https://aclanthology.org/2024.lrec-main.377} {Culturax: {A} cleaned, enormous, and multilingual dataset for large language models in 167 languages}.
\newblock In \emph{Proceedings of the 2024 Joint International Conference on Computational Linguistics, Language Resources and Evaluation, {LREC/COLING} 2024, 20-25 May, 2024, Torino, Italy}, pages 4226--4237. {ELRA} and {ICCL}.

\bibitem[{Oren et~al.(2019)Oren, Sagawa, Hashimoto, and Liang}]{DistributionallyRobust_OSHL19}
Yonatan Oren, Shiori Sagawa, Tatsunori~B. Hashimoto, and Percy Liang. 2019.
\newblock \href {https://doi.org/10.18653/v1/D19-1432} {Distributionally robust language modeling}.
\newblock In \emph{Proceedings of the 2019 Conference on Empirical Methods in Natural Language Processing and the 9th International Joint Conference on Natural Language Processing (EMNLP-IJCNLP)}, pages 4227--4237, Hong Kong, China. Association for Computational Linguistics.

\bibitem[{Paperno et~al.(2016)Paperno, Kruszewski, Lazaridou, Pham, Bernardi, Pezzelle, Baroni, Boleda, and Fern{\'a}ndez}]{LAMBADADataset_PKL+16}
Denis Paperno, Germ{\'a}n Kruszewski, Angeliki Lazaridou, Ngoc~Quan Pham, Raffaella Bernardi, Sandro Pezzelle, Marco Baroni, Gemma Boleda, and Raquel Fern{\'a}ndez. 2016.
\newblock \href {https://doi.org/10.18653/v1/P16-1144} {The {LAMBADA} dataset: Word prediction requiring a broad discourse context}.
\newblock In \emph{Proceedings of the 54th Annual Meeting of the Association for Computational Linguistics (Volume 1: Long Papers)}, pages 1525--1534, Berlin, Germany. Association for Computational Linguistics.

\bibitem[{Parmar et~al.(2024)Parmar, Prabhumoye, Jennings, Patwary, Subramanian, Su, Zhu, Narayanan, Jhunjhunwala, Dattagupta, Jawa, Liu, Mahabaleshwarkar, Nitski, Brundyn, Maki, Martinez, You, Kamalu, LeGresley, Fridman, Casper, Aithal, Kuchaiev, Shoeybi, Cohen, and Catanzaro}]{Nemotron415B_PPJ+24}
Jupinder Parmar, Shrimai Prabhumoye, Joseph Jennings, Mostofa Patwary, Sandeep Subramanian, Dan Su, Chen Zhu, Deepak Narayanan, Aastha Jhunjhunwala, Ayush Dattagupta, Vibhu Jawa, Jiwei Liu, Ameya Mahabaleshwarkar, Osvald Nitski, Annika Brundyn, James Maki, Miguel Martinez, Jiaxuan You, John Kamalu, Patrick LeGresley, Denys Fridman, Jared Casper, Ashwath Aithal, Oleksii Kuchaiev, Mohammad Shoeybi, Jonathan Cohen, and Bryan Catanzaro. 2024.
\newblock \href {https://arxiv.org/abs/2402.16819} {Nemotron-4 {{15B Technical Report}}}.

\bibitem[{Rajpurkar et~al.(2018)Rajpurkar, Jia, and Liang}]{KnowWhat_RJL18}
Pranav Rajpurkar, Robin Jia, and Percy Liang. 2018.
\newblock \href {https://doi.org/10.18653/v1/P18-2124} {Know what you don{'}t know: Unanswerable questions for {SQ}u{AD}}.
\newblock In \emph{Proceedings of the 56th Annual Meeting of the Association for Computational Linguistics (Volume 2: Short Papers)}, pages 784--789, Melbourne, Australia. Association for Computational Linguistics.

\bibitem[{Rao et~al.(2024)Rao, Liu, Lian, Cheng, Liao, and Zhang}]{commonIT}
Jun Rao, Xuebo Liu, Lian Lian, Shengjun Cheng, Yunjie Liao, and Min Zhang. 2024.
\newblock \href {https://doi.org/10.18653/v1/2024.emnlp-main.561} {{C}ommon{IT}: Commonality-aware instruction tuning for large language models via data partitions}.
\newblock In \emph{EMNLP}, pages 10064--10083, Miami, Florida, USA. Association for Computational Linguistics.

\bibitem[{Rao et~al.(2023)Rao, Meng, Ding, Qi, Liu, Zhang, and Tao}]{pesf-kd}
Jun Rao, Xv~Meng, Liang Ding, Shuhan Qi, Xuebo Liu, Min Zhang, and Dacheng Tao. 2023.
\newblock \href {https://doi.org/10.1109/TMM.2023.3321480} {Parameter-efficient and student-friendly knowledge distillation}.
\newblock \emph{{IEEE} Trans. Multim.}, pages 1--12.

\bibitem[{Sagawa et~al.(2019)Sagawa, Koh, Hashimoto, and Liang}]{DistributionallyRobust_SKHL20}
Shiori Sagawa, Pang~Wei Koh, Tatsunori~B. Hashimoto, and Percy Liang. 2019.
\newblock \href {https://arxiv.org/abs/1911.08731} {Distributionally {{Robust Neural Networks}} for {{Group Shifts}}: {{On}} the {{Importance}} of {{Regularization}} for {{Worst-Case Generalization}}}.

\bibitem[{Sakaguchi et~al.(2020)Sakaguchi, Bras, Bhagavatula, and Choi}]{WinoGrandeAdversarial_SBBC19}
Keisuke Sakaguchi, Ronan~Le Bras, Chandra Bhagavatula, and Yejin Choi. 2020.
\newblock \href {https://doi.org/10.1609/AAAI.V34I05.6399} {Winogrande: An adversarial winograd schema challenge at scale}.
\newblock In \emph{The Thirty-Fourth {AAAI} Conference on Artificial Intelligence, {AAAI} 2020, The Thirty-Second Innovative Applications of Artificial Intelligence Conference, {IAAI} 2020, The Tenth {AAAI} Symposium on Educational Advances in Artificial Intelligence, {EAAI} 2020, New York, NY, USA, February 7-12, 2020}, pages 8732--8740. {AAAI} Press.

\bibitem[{Shen et~al.(2023)Shen, Tao, Ma, Neiswanger, Liu, Wang, Tan, Hestness, Vassilieva, Soboleva, and Xing}]{SlimPajamaDCUnderstanding_STM+24}
Zhiqiang Shen, Tianhua Tao, Liqun Ma, Willie Neiswanger, Zhengzhong Liu, Hongyi Wang, Bowen Tan, Joel Hestness, Natalia Vassilieva, Daria Soboleva, and Eric Xing. 2023.
\newblock \href {https://arxiv.org/abs/2309.10818} {{{SlimPajama-DC}}: {{Understanding Data Combinations}} for {{LLM Training}}}.

\bibitem[{Shliazhko et~al.(2024)Shliazhko, Fenogenova, Tikhonova, Kozlova, Mikhailov, and Shavrina}]{MGPTFewShot_SFT+23}
Oleh Shliazhko, Alena Fenogenova, Maria Tikhonova, Anastasia Kozlova, Vladislav Mikhailov, and Tatiana Shavrina. 2024.
\newblock \href {https://doi.org/10.1162/TACL\_A\_00633} {mgpt: Few-shot learners go multilingual}.
\newblock \emph{Trans. Assoc. Comput. Linguistics}, 12:58--79.

\bibitem[{Sun et~al.(2024)Sun, Liu, Bair, and Kolter}]{SimpleEffective_SLBK23}
Mingjie Sun, Zhuang Liu, Anna Bair, and J.~Zico Kolter. 2024.
\newblock \href {https://openreview.net/forum?id=PxoFut3dWW} {A simple and effective pruning approach for large language models}.
\newblock In \emph{The Twelfth International Conference on Learning Representations, {ICLR} 2024, Vienna, Austria, May 7-11, 2024}. OpenReview.net.

\bibitem[{Tatman(2017)}]{GenderDialect_Tat17}
Rachael Tatman. 2017.
\newblock \href {https://doi.org/10.18653/v1/W17-1606} {Gender and dialect bias in {Y}ou{T}ube{'}s automatic captions}.
\newblock In \emph{Proceedings of the First {ACL} Workshop on Ethics in Natural Language Processing}, pages 53--59, Valencia, Spain. Association for Computational Linguistics.

\bibitem[{Tikhonov and Ryabinin(2021)}]{ItsAll_TR21}
Alexey Tikhonov and Max Ryabinin. 2021.
\newblock \href {https://doi.org/10.18653/v1/2021.findings-acl.310} {{I}t{'}s {A}ll in the {H}eads: {U}sing {A}ttention {H}eads as a {B}aseline for {C}ross-{L}ingual {T}ransfer in {C}ommonsense {R}easoning}.
\newblock In \emph{Findings of the Association for Computational Linguistics: ACL-IJCNLP 2021}, pages 3534--3546, Online. Association for Computational Linguistics.

\bibitem[{Touvron et~al.(2023{\natexlab{a}})Touvron, Lavril, Izacard, Martinet, Lachaux, Lacroix, Rozi{\`e}re, Goyal, Hambro, Azhar, Rodriguez, Joulin, Grave, and Lample}]{LlamaOpen_TLI+23}
Hugo Touvron, Thibaut Lavril, Gautier Izacard, Xavier Martinet, Marie-Anne Lachaux, Timoth{\'e}e Lacroix, Baptiste Rozi{\`e}re, Naman Goyal, Eric Hambro, Faisal Azhar, Aurelien Rodriguez, Armand Joulin, Edouard Grave, and Guillaume Lample. 2023{\natexlab{a}}.
\newblock \href {https://arxiv.org/abs/2302.13971} {{{LLaMA}}: {{Open}} and {{Efficient Foundation Language Models}}}.

\bibitem[{Touvron et~al.(2023{\natexlab{b}})Touvron, Martin, Stone, Albert, Almahairi, Babaei, Bashlykov, Batra, Bhargava, Bhosale, Bikel, Blecher, Ferrer, Chen, Cucurull, Esiobu, Fernandes, Fu, Fu, Fuller, Gao, Goswami, Goyal, Hartshorn, Hosseini, Hou, Inan, Kardas, Kerkez, Khabsa, Kloumann, Korenev, Koura, Lachaux, Lavril, Lee, Liskovich, Lu, Mao, Martinet, Mihaylov, Mishra, Molybog, Nie, Poulton, Reizenstein, Rungta, Saladi, Schelten, Silva, Smith, Subramanian, Tan, Tang, Taylor, Williams, Kuan, Xu, Yan, Zarov, Zhang, Fan, Kambadur, Narang, Rodriguez, Stojnic, Edunov, and Scialom}]{LlamaOpen_TMS+23}
Hugo Touvron, Louis Martin, Kevin Stone, Peter Albert, Amjad Almahairi, Yasmine Babaei, Nikolay Bashlykov, Soumya Batra, Prajjwal Bhargava, Shruti Bhosale, Dan Bikel, Lukas Blecher, Cristian~Canton Ferrer, Moya Chen, Guillem Cucurull, David Esiobu, Jude Fernandes, Jeremy Fu, Wenyin Fu, Brian Fuller, Cynthia Gao, Vedanuj Goswami, Naman Goyal, Anthony Hartshorn, Saghar Hosseini, Rui Hou, Hakan Inan, Marcin Kardas, Viktor Kerkez, Madian Khabsa, Isabel Kloumann, Artem Korenev, Punit~Singh Koura, Marie-Anne Lachaux, Thibaut Lavril, Jenya Lee, Diana Liskovich, Yinghai Lu, Yuning Mao, Xavier Martinet, Todor Mihaylov, Pushkar Mishra, Igor Molybog, Yixin Nie, Andrew Poulton, Jeremy Reizenstein, Rashi Rungta, Kalyan Saladi, Alan Schelten, Ruan Silva, Eric~Michael Smith, Ranjan Subramanian, Xiaoqing~Ellen Tan, Binh Tang, Ross Taylor, Adina Williams, Jian~Xiang Kuan, Puxin Xu, Zheng Yan, Iliyan Zarov, Yuchen Zhang, Angela Fan, Melanie Kambadur, Sharan Narang, Aurelien Rodriguez, Robert Stojnic, Sergey Edunov, and Thomas
  Scialom. 2023{\natexlab{b}}.
\newblock \href {https://arxiv.org/abs/2307.09288} {Llama 2: {{Open Foundation}} and {{Fine-Tuned Chat Models}}}.

\bibitem[{Wang et~al.(2020{\natexlab{a}})Wang, Zhang, and Grosse}]{PickingWinning_WZG20}
Chaoqi Wang, Guodong Zhang, and Roger~B. Grosse. 2020{\natexlab{a}}.
\newblock \href {https://openreview.net/forum?id=SkgsACVKPH} {Picking winning tickets before training by preserving gradient flow}.
\newblock In \emph{8th International Conference on Learning Representations, {ICLR} 2020, Addis Ababa, Ethiopia, April 26-30, 2020}. OpenReview.net.

\bibitem[{Wang et~al.(2024)Wang, Li, Chen, Cai, Zhu, Lin, Cao, Kong, Liu, Liu, and Sui}]{LargeLanguage_WLC+24}
Peiyi Wang, Lei Li, Liang Chen, Zefan Cai, Dawei Zhu, Binghuai Lin, Yunbo Cao, Lingpeng Kong, Qi~Liu, Tianyu Liu, and Zhifang Sui. 2024.
\newblock \href {https://doi.org/10.18653/v1/2024.acl-long.511} {Large language models are not fair evaluators}.
\newblock In \emph{Proceedings of the 62nd Annual Meeting of the Association for Computational Linguistics (Volume 1: Long Papers)}, pages 9440--9450, Bangkok, Thailand. Association for Computational Linguistics.

\bibitem[{Wang et~al.(2020{\natexlab{b}})Wang, Wei, Dong, Bao, Yang, and Zhou}]{DBLP:conf/nips/WangW0B0020}
Wenhui Wang, Furu Wei, Li~Dong, Hangbo Bao, Nan Yang, and Ming Zhou. 2020{\natexlab{b}}.
\newblock \href {https://proceedings.neurips.cc/paper/2020/hash/3f5ee243547dee91fbd053c1c4a845aa-Abstract.html} {Minilm: Deep self-attention distillation for task-agnostic compression of pre-trained transformers}.
\newblock In \emph{Advances in Neural Information Processing Systems 33: Annual Conference on Neural Information Processing Systems 2020, NeurIPS 2020, December 6-12, 2020, virtual}.

\bibitem[{Wang et~al.(2020{\natexlab{c}})Wang, Tsvetkov, and Neubig}]{BalancingTraining_WTN20}
Xinyi Wang, Yulia Tsvetkov, and Graham Neubig. 2020{\natexlab{c}}.
\newblock \href {https://doi.org/10.18653/v1/2020.acl-main.754} {Balancing training for multilingual neural machine translation}.
\newblock In \emph{Proceedings of the 58th Annual Meeting of the Association for Computational Linguistics}, pages 8526--8537, Online. Association for Computational Linguistics.

\bibitem[{Welbl et~al.(2017)Welbl, Liu, and Gardner}]{CrowdsourcingMultiple_WLG17}
Johannes Welbl, Nelson~F. Liu, and Matt Gardner. 2017.
\newblock \href {https://doi.org/10.18653/v1/W17-4413} {Crowdsourcing multiple choice science questions}.
\newblock In \emph{Proceedings of the 3rd Workshop on Noisy User-generated Text}, pages 94--106, Copenhagen, Denmark. Association for Computational Linguistics.

\bibitem[{Wen et~al.(2016)Wen, Wu, Wang, Chen, and Li}]{LearningStructured_WWW+16}
Wei Wen, Chunpeng Wu, Yandan Wang, Yiran Chen, and Hai Li. 2016.
\newblock \href {https://proceedings.neurips.cc/paper/2016/hash/41bfd20a38bb1b0bec75acf0845530a7-Abstract.html} {Learning structured sparsity in deep neural networks}.
\newblock In \emph{Advances in Neural Information Processing Systems 29: Annual Conference on Neural Information Processing Systems 2016, December 5-10, 2016, Barcelona, Spain}, pages 2074--2082.

\bibitem[{Xia et~al.(2024)Xia, Gao, Zeng, and Chen}]{ShearedLLaMA_XGZC23}
Mengzhou Xia, Tianyu Gao, Zhiyuan Zeng, and Danqi Chen. 2024.
\newblock \href {https://openreview.net/forum?id=09iOdaeOzp} {Sheared llama: Accelerating language model pre-training via structured pruning}.
\newblock In \emph{The Twelfth International Conference on Learning Representations, {ICLR} 2024, Vienna, Austria, May 7-11, 2024}. OpenReview.net.

\bibitem[{Xia et~al.(2022)Xia, Zhong, and Chen}]{StructuredPruning_XZC22}
Mengzhou Xia, Zexuan Zhong, and Danqi Chen. 2022.
\newblock \href {https://doi.org/10.18653/v1/2022.acl-long.107} {Structured pruning learns compact and accurate models}.
\newblock In \emph{Proceedings of the 60th Annual Meeting of the Association for Computational Linguistics (Volume 1: Long Papers)}, pages 1513--1528, Dublin, Ireland. Association for Computational Linguistics.

\bibitem[{Yang et~al.(2024)Yang, Yang, Hui, Zheng, Yu, Zhou, Li, Li, Liu, Huang, Dong, Wei, Lin, Tang, Wang, Yang, Tu, Zhang, Ma, Yang, Xu, Zhou, Bai, He, Lin, Dang, Lu, Chen, Yang, Li, Xue, Ni, Zhang, Wang, Peng, Men, Gao, Lin, Wang, Bai, Tan, Zhu, Li, Liu, Ge, Deng, Zhou, Ren, Zhang, Wei, Ren, Liu, Fan, Yao, Zhang, Wan, Chu, Liu, Cui, Zhang, Guo, and Fan}]{Qwen2Technical_YYH+24}
An~Yang, Baosong Yang, Binyuan Hui, Bo~Zheng, Bowen Yu, Chang Zhou, Chengpeng Li, Chengyuan Li, Dayiheng Liu, Fei Huang, Guanting Dong, Haoran Wei, Huan Lin, Jialong Tang, Jialin Wang, Jian Yang, Jianhong Tu, Jianwei Zhang, Jianxin Ma, Jianxin Yang, Jin Xu, Jingren Zhou, Jinze Bai, Jinzheng He, Junyang Lin, Kai Dang, Keming Lu, Keqin Chen, Kexin Yang, Mei Li, Mingfeng Xue, Na~Ni, Pei Zhang, Peng Wang, Ru~Peng, Rui Men, Ruize Gao, Runji Lin, Shijie Wang, Shuai Bai, Sinan Tan, Tianhang Zhu, Tianhao Li, Tianyu Liu, Wenbin Ge, Xiaodong Deng, Xiaohuan Zhou, Xingzhang Ren, Xinyu Zhang, Xipin Wei, Xuancheng Ren, Xuejing Liu, Yang Fan, Yang Yao, Yichang Zhang, Yu~Wan, Yunfei Chu, Yuqiong Liu, Zeyu Cui, Zhenru Zhang, Zhifang Guo, and Zhihao Fan. 2024.
\newblock \href {https://arxiv.org/abs/2407.10671} {Qwen2 {{Technical Report}}}.

\bibitem[{Yang et~al.(2019)Yang, Zhang, Tar, and Baldridge}]{PAWSXCrosslingual_YZTB19}
Yinfei Yang, Yuan Zhang, Chris Tar, and Jason Baldridge. 2019.
\newblock \href {https://doi.org/10.18653/v1/D19-1382} {{PAWS}-{X}: A cross-lingual adversarial dataset for paraphrase identification}.
\newblock In \emph{Proceedings of the 2019 Conference on Empirical Methods in Natural Language Processing and the 9th International Joint Conference on Natural Language Processing (EMNLP-IJCNLP)}, pages 3687--3692, Hong Kong, China. Association for Computational Linguistics.

\bibitem[{Zellers et~al.(2019)Zellers, Holtzman, Bisk, Farhadi, and Choi}]{HellaSwagCan_ZHB+19}
Rowan Zellers, Ari Holtzman, Yonatan Bisk, Ali Farhadi, and Yejin Choi. 2019.
\newblock \href {https://doi.org/10.18653/v1/P19-1472} {{H}ella{S}wag: Can a machine really finish your sentence?}
\newblock In \emph{Proceedings of the 57th Annual Meeting of the Association for Computational Linguistics}, pages 4791--4800, Florence, Italy. Association for Computational Linguistics.

\bibitem[{Zhang et~al.(2024{\natexlab{a}})Zhang, Liu, Cherry, and Firat}]{WhenScaling_ZLCF24}
Biao Zhang, Zhongtao Liu, Colin Cherry, and Orhan Firat. 2024{\natexlab{a}}.
\newblock \href {https://openreview.net/forum?id=5HCnKDeTws} {When scaling meets {LLM} finetuning: The effect of data, model and finetuning method}.
\newblock In \emph{The Twelfth International Conference on Learning Representations, {ICLR} 2024, Vienna, Austria, May 7-11, 2024}. OpenReview.net.

\bibitem[{Zhang et~al.(2024{\natexlab{b}})Zhang, Chen, Shen, Yang, Ou, Yu, and Zhuang}]{LoRAPruneStructured_ZCS+24}
Mingyang Zhang, Hao Chen, Chunhua Shen, Zhen Yang, Linlin Ou, Xinyi Yu, and Bohan Zhuang. 2024{\natexlab{b}}.
\newblock \href {https://doi.org/10.18653/v1/2024.findings-acl.178} {{L}o{RAP}rune: Structured pruning meets low-rank parameter-efficient fine-tuning}.
\newblock In \emph{Findings of the Association for Computational Linguistics ACL 2024}, pages 3013--3026, Bangkok, Thailand and virtual meeting. Association for Computational Linguistics.

\bibitem[{Zhao et~al.(2023)Zhao, Gu, Varma, Luo, Huang, Xu, Wright, Shojanazeri, Ott, Shleifer, Desmaison, Balioglu, Damania, Nguyen, Chauhan, Hao, Mathews, and Li}]{PyTorchFSDP_ZGV+23}
Yanli Zhao, Andrew Gu, Rohan Varma, Liang Luo, Chien-Chin Huang, Min Xu, Less Wright, Hamid Shojanazeri, Myle Ott, Sam Shleifer, Alban Desmaison, Can Balioglu, Pritam Damania, Bernard Nguyen, Geeta Chauhan, Yuchen Hao, Ajit Mathews, and Shen Li. 2023.
\newblock \href {https://arxiv.org/abs/2304.11277} {{{PyTorch FSDP}}: {{Experiences}} on {{Scaling Fully Sharded Data Parallel}}}.

\bibitem[{Zhou et~al.(2021)Zhou, Levy, Li, Ghazvininejad, and Neubig}]{DistributionallyRobust_ZLL+21}
Chunting Zhou, Daniel Levy, Xian Li, Marjan Ghazvininejad, and Graham Neubig. 2021.
\newblock \href {https://doi.org/10.18653/v1/2021.emnlp-main.458} {Distributionally robust multilingual machine translation}.
\newblock In \emph{Proceedings of the 2021 Conference on Empirical Methods in Natural Language Processing}, pages 5664--5674, Online and Punta Cana, Dominican Republic. Association for Computational Linguistics.

\end{thebibliography}
\appendix
\newpage
\
\newpage

\newcommand{\dashline}[1]{
  \noalign{\vskip 0.1ex}
  \cdashline{#1}
  \noalign{\vskip 0.3ex}
}

\section{Detailed Experimental Setup}

\begin{table*}[!ht]
 \centering
 \small
 \begin{tabular}{lcc} \toprule
 & \bf Pruning & \bf Contined Pretraining \\ \midrule
 \bf Training Steps & 3,200 & 48,000 \\
 \bf Learning rate of $z, \phi, \lambda$ & $1.0$ & - \\
 \bf Learning Rate of $\theta$ & $0.0001$ & $0.0001$ \\
 \bf LR warmup ratio & $10\%$ & $3\%$ \\
 \bf Batch size (tokens) & $131$K & $1$M \\
 \bf Ratio update interval $m$ (steps) & $50$ & $400$ \\ 
 \bottomrule 
 \end{tabular}
 \caption{\label{tab:setting_main}
 Training hyperparameters for the main experiment.}
\end{table*}

\begin{table*}[!ht]
 \centering
 \small
 \begin{tabular}{lcccccc}
 \toprule
 \textbf{Model} & \textbf{\#Param} & \textbf{\#Layers} & \textbf{Hidden} & \textbf{Intermediate} & \textbf{\#Heads} & \textbf{Head Dim}\\
 \midrule
 \bf Pruned-0.5B & 0.5B & 24 & 1024 & 2816 & 8 & 128 \\
 \bf Pruned-1.3B & 1.3B & 24 & 2048 & 5504 & 16 & 128 \\
 \bf Pruned-2.7B & 2.7B & 32 & 2560 & 6912 & 20 & 128 \\
 \midrule
 \bf Llama2-7B & 6.7B & 32 & 4096 & 11008 & 32 & 128 \\ 
 \bottomrule
 \end{tabular}
 \caption{\label{tab:setting_mainmodel}
 The model configurations for the target model of pruning and the base models for the main experiment.}
\end{table*}

\subsection{Main Experiment}
\label{apx:settingMain}

\paragraph{Model training.}
All experiments are conducted on 8 NVIDIA A100 40GB GPUs. The training hyperparameters for the main experiment are listed in Table~\ref{tab:setting_main}, and the target model configuration for pruning is detailed in Table~\ref{tab:setting_mainmodel}. Pruning takes approximately 12 hours for both the 1.3B and 2.7B models. Continued pretraining requires around 9 days for the 1.3B model and 18 days for the 2.7B model.

For both pruning and continued pretraining, we follow the configurations of Sheared Llama as closely as possible. We use fully sharded data parallel \citep{PyTorchFSDP_ZGV+23} for parallel training and FlashAttention V1 \citep{FlashAttentionFast_DFE+22} to speed up the training process. A cosine learning rate scheduler is employed, reducing the learning rate to 10\% of its peak value.

\paragraph{DRO.}
We follow Sheared Llama to update the data ratio every 50 steps during pruning and every 400 steps during continued pretraining. For the DRO setup, we follow \citet{DistributionallyRobust_ZLL+21} closely. The constraint size $\rho$ for the chi-square ball is set to \{0.05, 0.1, 0.2\}. Preliminary experiments show that $\rho = 0.1$ yields the best results, so we use this value in all experiments. Following their setup, we truncate the dynamic data ratio to prevent it from dropping below the minimum reference data ratio, which further ensures balanced domain training. We compute historical loss values using an exponential moving average, with the hyperparameter $\lambda$ set to 0.1, which is also used for updating the reference data ratio. Besides, for the prediction of the reference loss, we maintain an average loss below $3 \times 10^{-5}$, demonstrating the effectiveness of our method.

\paragraph{Instruction tuning.}
For instruction tuning, the instruction begins with "You are a helpful assistant. Write a response that appropriately completes the request." We perform full-parameter fine-tuning for 5 epochs, with a learning rate of $5e^{-5}$, a warmup ratio of $3\%$, and a batch size of $128$.

To evaluate instruction tuning, we follow the methodology of Sheared Llama, using LLMs to assess model performance. Given outputs from two models, we ask the LLM to determine which is better using the prompt: ``Here is the user request: .... Here are the two outputs for this request: Output A: .... Output B: .... Which output is better, A or B?''. Since \citet{LargeLanguage_WLC+24} note that using GPT models as evaluators can lead to preference shifts when output order is reversed, we randomly switch the positions of the outputs to ensure each result appears as Output A or Output B equally. We report the average win rate to mitigate position bias. The model \texttt{gpt-4o-2024-08-06} is used for evaluation.

\begin{table*}[!ht]
 \centering
 \small
 \begin{tabular}{lccccccc}
 \toprule
 \textbf{Model} & \textbf{\#Param} & \textbf{\#Layers} & \textbf{Hidden} & \textbf{Intermediate} & \textbf{\#Heads} & \textbf{\#KV Heads} & \textbf{Head Dim}\\
 \midrule
 \bf Pruned-1.8B & 1.8B & 28 & 1536 & 8960 & 14 & 2 & 128 \\
 \midrule
 \bf Qwen2-1.5B & 1.5B & 28 & 1536 & 8960 & 12 & 2 & 128 \\ 
 \bf Qwen2-7B & 7.6B & 28 & 3584 & 18944 & 28 & 4 & 128 \\ 
 \bottomrule
 \end{tabular}
 \caption{\label{tab:setting_multimodel}
 The model configurations for the target model of pruning and the base models for the multilingual experiment. }
\end{table*}

\subsection{Multilingual Experiment}
\label{apx:settingMultilingual}

\paragraph{Model training.} 
The experimental setup is largely consistent with Appendix~\ref{apx:settingMain}. Given the extended training duration, we standardize the training to 40,000 steps. The configuration of the target model for pruning is detailed in Table~\ref{tab:setting_multimodel}. To maintain a consistent ratio between the number of heads and KV heads required by the structured pruning method, we keep the head dimension unchanged and add two extra heads, increasing the number of parameters to 1.8B. Continued pretraining takes around 8 days for the 1.5B model and 12 days for the 1.8B model.

\paragraph{Data.} 
We use the CulturaX dataset \citep{CulturaXCleaned_NVL+23}, a large multilingual resource with 6.3 trillion tokens across 167 languages, integrating mC4 and OSCAR, and meticulously cleaned and deduplicated. We select eight languages covered by Qwen2: English (EN), Russian (RU), Chinese (ZH), Japanese (JA), Arabic (AR), Turkish (TR), Korean (KO), and Thai (TH), representing diverse language families.

\paragraph{Metrics.}
We adopt the experimental setups from previous studies and evaluate performance on downstream tasks in a zero-shot setting. Specifically, we follow XGLM \citep{FewshotLearning_LMA+22} and mGPT \citep{MGPTFewShot_SFT+23}, covering tasks such as natural language inference (XNLI; \citealp{XNLIEvaluating_CRL+18}), Winograd schema challenge (XWINO; \citealp{ItsAll_TR21}), commonsense reasoning (XStoryCloze; \citealp{FewshotLearning_LMA+22}), and paraphrase detection (PAWSX; \citealp{PAWSXCrosslingual_YZTB19}). Task coverage varies across languages, and not all tasks include all languages in our training set. We report results for languages overlapping between tasks and our training set, providing average performance if a language appears in multiple tasks. The lm-evaluation-harness package \citep{eval-harness} is used for the comprehensive evaluation of downstream tasks.

\section{Supplement Experimental Results}

\subsection{Analysis of Mask Similarity}
\label{apx:analysisMask}

\begin{figure}[!ht]
 \centering
 \includegraphics[width=\linewidth]{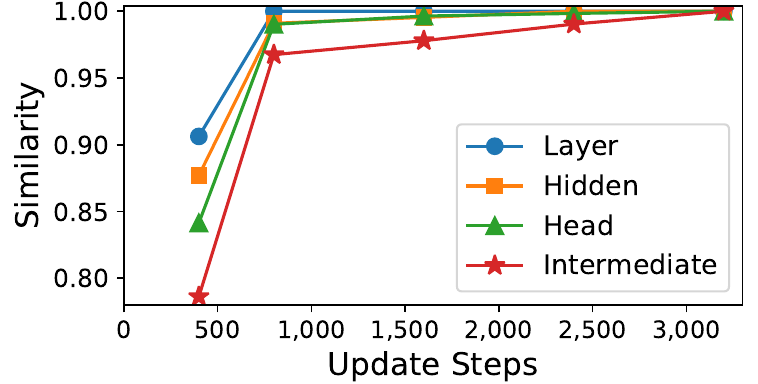}
 \caption{\label{fig:mask_converge}
 Convergence of masks over 3200 pruning steps. Similarity indicates the similarity between pruning decisions at a certain step and the final decisions at step 3200. ``Layer'', ``Hidden'', ``Head'', and ``Intermediate'' correspond to the four pruning dimensions.}
 \vspace{6pt}
\end{figure}

\begin{table}[!ht]
\small
\centering
\begin{tabular}{lcc}
\toprule
\bf Structure & \bf Same & \bf Different \\ \midrule
\bf Layer & 81.25$\pm$3.61 & 78.65$\pm$2.24 \\
\bf Hidden & 60.33$\pm$0.78 & 65.02$\pm$1.47 \\
\bf Head & 68.88$\pm$0.47 & 70.53$\pm$0.65 \\
\bf Intermediate & 56.37$\pm$0.15 & 56.40$\pm$0.14 \\ \bottomrule
\end{tabular}
\caption{\label{tab:mask_sim}
Mask similarity mean values and the standard error of the mean under different data scheduling strategies and random seeds. ``Same'' indicates using identical data scheduling but different random seeds, while ``Different'' indicates using different data scheduling strategies.
}
\end{table}

\begin{table*}[!t]
\small
\centering
\scalebox{0.95}{
\begin{tabular}{lcccccccc}
\toprule
\multirow{3}[3]{*}{\bf Tasks} & \multicolumn{4}{c}{\bf Pruning} & \multicolumn{4}{c}{\bf Continued Pretraining} \\ \cmidrule(lr){2-5}\cmidrule(lr){6-9}
~ & \multicolumn{2}{c}{\bf To: 1.3B} & \multicolumn{2}{c}{\bf To: 2.7B} & \multicolumn{2}{c}{\bf To: 1.3B} & \multicolumn{2}{c}{\bf To: 2.7B} \\ \cmidrule(lr){2-3}\cmidrule(lr){4-5}\cmidrule(lr){6-7}\cmidrule(lr){8-9}
~ & \bf From: 7B & \bf From: 13B & \bf From: 7B & \bf From: 13B & \bf From: 7B & \bf From: 13B & \bf From: 7B & \bf From: 13B \\
\midrule
\bf ARCC (25) & \bf 23.21 & 22.10 & \bf 30.29 & 27.30 & \bf 33.62 & 32.17 & \bf 40.53 & 40.36 \\
\bf ARCE & \bf 42.26 & 40.07 & \bf 53.11 & 47.31 & \bf 60.90 & 58.92 & \bf 67.13 & 66.58 \\
\bf BoolQ & 59.69 & \bf 59.88 & 59.36 & \bf 60.12 & \bf 63.36 & 56.88 & 65.08 & \bf 67.13 \\
\bf HelS (10) & \bf 35.27 & 32.38 & \bf 48.07 & 41.62 & \bf 58.88 & 58.70 & \bf 69.22 & 67.67 \\
\bf LAMB & \bf 38.27 & 34.08 & \bf 51.80 & 46.56 & \bf 60.28 & 59.87 & \bf 66.91 & 66.23 \\
\bf LogiQA & \bf 26.73 & 25.50 & \bf 27.19 & 24.42 & \bf 28.88 & 25.65 & 28.73 & \bf 29.80 \\
\bf MMLU (5) & 24.82 & \bf 25.78 & 24.86 & \bf 25.56 & \bf 27.28 & 26.76 & 26.99 & \bf 27.60 \\
\bf NQ (5) & \bf 1.91 & 1.52 & \bf 4.02 & 2.35 & \bf 10.44 & 8.86 & \bf 15.82 & 13.24 \\
\bf PIQA & \bf 61.81 & 61.32 & \bf 67.08 & 64.15 & \bf 72.69 & 72.31 & \bf 75.19 & 74.27 \\
\bf SciQ & \bf 79.80 & 79.60 & \bf 86.30 & 83.30 & \bf 87.70 & 87.30 & 89.80 & \bf 91.00 \\
\bf SQuAD & \bf 13.80 & 6.87 & 17.05 & \bf 18.61 & \bf 35.06 & 28.52 & 44.69 & \bf 46.55 \\
\bf TriQA (5) & \bf 5.26 & 3.33 & \bf 11.75 & 6.77 & \bf 28.10 & 24.40 & \bf 43.33 & 38.17 \\
\bf TruthQA & 32.99 & \bf 34.15 & 31.12 & \bf 31.50 & \bf 29.68 & 29.66 & 30.13 & \bf 30.75 \\
\bf WinoG & \bf 51.62 & 49.09 & \bf 54.14 & 53.83 & \bf 58.01 & 56.83 & \bf 64.72 & 62.04 \\
\bf WSC & \bf 36.54 & \bf 36.54 & \bf 36.54 & \bf 36.54 & 50.00 & \bf 60.58 & \bf 46.15 & 36.54 \\ \midrule
\bf Average & \bf 35.60 & 34.15 & \bf 40.18 & 38.00 & \bf 46.99 & 45.83 & \bf 51.63 & 50.53 \\
\bottomrule
\end{tabular}}
\caption{\label{tab:pruneLarger}
The performance of pruning Llama2-7B and 13B models down to 1.3B and 2.7B parameters. ``Pruning'' refers to using the pruned model without continued pretraining, while ``Continue Pretraining'' means using the model after continued pretraining. ``From'' and ``To'' indicate the size of the source and target model, respectively. \textbf{Bold} indicates superior performance when pruning from 7B or 13B.}
\end{table*}

To perform a detailed analysis of the masks, we extract the masks generated during training and prune the model by removing components with the lowest scores, shaping the model according to the target specifications. We then calculate the probability of the model making consistent pruning decisions for each substructure and examine how different training steps or strategies influence the masks. Specifically, we apply Sheared Llama, constant, and our proposed strategies, using two distinct random seeds for pruning, which yields six unique 1.3B models. We then analyze the similarities across these models.

\paragraph{The masks converge quickly during training.}
The convergence speed of the masks during training is illustrated in Figure~\ref{fig:mask_converge}. Pruning achieves over 75\% similarity within 400 steps and over 95\% within 800 steps, indicating that effective results can be obtained with relatively few pruning steps. While layer pruning converges rapidly, pruning intermediates of fully connected layers is slower, suggesting that coarser-grained decisions converge more quickly than finer-grained decisions.

\paragraph{The randomness of pruning decisions is significant.}
We analyze mask similarity across training sessions with different random seeds under identical and distinct data scheduling strategies. The results, shown in Table~\ref{tab:mask_sim}, present mean values and standard error of the mean. The trend across pruning dimensions aligns with the previous findings: similarity is higher at the coarse-grained layer level and lower at the finer-grained intermediate level. Besides, comparisons across three pairs under identical settings and twelve pairs under different settings show consistently low similarity. However, no significant difference in perplexity is observed, suggesting that the model's interchangeable parameters allow similar outcomes despite different pruning decisions. This randomness obscures variations caused by differing data distributions.

\subsection{Pruning from Larger LLMs}
\label{apx:largerLLM}

We investigate whether pruning from LLMs with a higher pruning ratio provides additional benefits. Experiments are conducted in the monolingual setting, consistent with the main text, to compare the effects of pruning from Llama2-7B and Llama2-13B.

The results, presented in Table~\ref{tab:pruneLarger}, indicate that pruning from the 13B model consistently yields worse outcomes, regardless of whether continued pretraining is applied. On average, this approach results in a downstream performance decrease of 1.47. These findings suggest that pruning from a larger model leads to a more significant performance decline, often producing inferior results under a fixed training budget, especially under a high pruning ratio.

\begin{table*}[!t]
\small
\centering
\begin{tabular}{lcccccccccccc}
\toprule
\multirow{2}[2]{*}{\bf Tasks} & \multicolumn{6}{c}{\bf ReSheared 1.3B} & \multicolumn{6}{c}{\bf DRPruning 1.3B} \\ \cmidrule(lr){2-7}\cmidrule(lr){8-13}
& \bf P1 & \bf P2 & \bf P3 & \bf P4 & \bf P5 & \bf Avg. & \bf P1 & \bf P2 & \bf P3 & \bf P4 & \bf P5 & \bf Avg. \\
\midrule
\bf ARCC (25) & \bf 34.30 & \bf 33.62 & \bf 34.04 & \bf 33.53 & \bf 35.15 & \bf 34.13 & 33.62 & 33.36 & 33.45 & 33.28 & 34.13 & 33.57 \\
\bf ARCE & 60.35 & \bf 59.76 & 61.03 & 59.05 & 60.98 & 60.23 & \bf 60.90 & 58.00 & \bf 61.53 & \bf 59.81 & \bf 61.28 & \bf 60.30 \\
\bf BoolQ & 61.01 & 58.90 & \bf 62.32 & \bf 62.26 & 62.48 & 61.38 & \bf 63.36 & \bf 63.12 & 61.50 & 59.11 & \bf 62.48 & \bf 61.93 \\
\bf HelS (10) & \bf 63.06 & \bf 63.05 & \bf 63.12 & \bf 62.99 & \bf 63.10 & \bf 63.06 & 58.88 & 58.77 & 58.66 & 58.66 & 58.82 & 58.76 \\
\bf LAMB & 58.84 & 59.83 & 59.65 & 60.02 & 60.02 & 59.67 & \bf 60.28 & \bf 60.90 & \bf 61.09 & \bf 61.28 & \bf 61.23 & \bf 60.96 \\
\bf LogiQA & 28.11 & 28.11 & \bf 31.03 & 28.42 & \bf 29.34 & 29.00 & \bf 28.88 & \bf 29.49 & 29.65 & \bf 29.19 & 27.80 & \bf 29.03 \\
\bf MMLU (5) & 26.60 & 25.92 & 25.69 & 25.56 & 25.61 & 25.87 & \bf 27.28 & \bf 26.79 & \bf 26.86 & \bf 26.54 & \bf 26.33 & \bf 26.76 \\
\bf NQ (5) & 8.39 & 7.73 & 8.31 & 7.98 & 8.34 & 8.14 & \bf 10.44 & \bf 9.58 & \bf 9.75 & \bf 9.11 & \bf 10.08 & \bf 9.79 \\
\bf PIQA & \bf 74.59 & \bf 74.43 & \bf 74.92 & \bf 73.88 & \bf 74.65 & \bf 74.49 & 72.69 & 71.98 & 72.09 & 72.20 & 72.47 & 72.27 \\
\bf SciQ & 86.40 & 85.70 & 88.20 & 85.90 & 87.50 & 86.76 & \bf 87.70 & \bf 88.30 & \bf 89.40 & \bf 88.20 & \bf 89.50 & \bf 88.64 \\
\bf SQuAD & \bf 37.59 & \bf 34.25 & \bf 40.39 & \bf 44.09 & \bf 35.76 & \bf 38.43 & 35.06 & 33.44 & 39.59 & 41.53 & 32.31 & 36.38 \\
\bf TriQA (5) & 24.98 & 25.06 & 25.10 & 23.61 & 25.00 & 24.75 & \bf 28.10 & \bf 27.62 & \bf 28.45 & \bf 26.47 & \bf 28.01 & \bf 27.72 \\
\bf TruthQA & 28.09 & 29.84 & 30.33 & 29.20 & 29.16 & 29.31 & \bf 29.68 & \bf 32.07 & \bf 31.69 & \bf 30.44 & \bf 30.87 & \bf 30.95 \\
\bf WinoG & \bf 60.06 & \bf 59.59 & 59.12 & \bf 61.01 & \bf 59.04 & \bf 59.68 & 58.01 & 59.27 & \bf 60.62 & 59.35 & 58.64 & 59.21 \\
\bf WSC & 40.38 & 36.54 & 36.54 & 36.54 & 41.35 & 38.27 & \bf 50.00 & \bf 50.00 & \bf 49.04 & \bf 55.77 & \bf 48.08 & \bf 50.58 \\ \midrule
\bf Average & 46.18 & 45.49 & 46.65 & 46.27 & 46.50 & 46.21 & \bf 46.99 & \bf 46.85 & \bf 47.56 & \bf 47.40 & \bf 46.80 & \bf 47.12 \\
\bottomrule
\end{tabular}
\caption{Performance comparison between ReSheared 1.3B and DRPruning 1.3B across five different prompts. ``P1'' to ``P5'' represent five distinct prompts. Other abbreviations follow the definitions in Table~\ref{tab:main}. \textbf{Bold} indicates superior performance when comparing ReSheared and DRPruning.}
\label{tab:comparison13}
\end{table*}
\begin{table*}[!t]
\small
\centering
\begin{tabular}{lcccccccccccc}
\toprule
\multirow{2}[2]{*}{\bf Tasks} & \multicolumn{6}{c}{\bf ReSheared 2.7B} & \multicolumn{6}{c}{\bf DRPruning 2.7B} \\ \cmidrule(lr){2-7}\cmidrule(lr){8-13}
& \bf P1 & \bf P2 & \bf P3 & \bf P4 & \bf P5 & \bf Avg. & \bf P1 & \bf P2 & \bf P3 & \bf P4 & \bf P5 & \bf Avg. \\
\midrule
\bf ARCC (25) & 40.10 & \bf 39.85 & 40.44 & 40.36 & 40.10 & 40.17 & \bf 40.53 & 39.08 & \bf 40.44 & \bf 40.96 & \bf 40.96 & \bf 40.39 \\
\bf ARCE & \bf 67.72 & \bf 67.30 & \bf 67.42 & 63.55 & 67.38 & \bf 66.67 & 67.13 & 64.52 & 67.26 & \bf 64.39 & \bf 67.55 & 66.14 \\
\bf BoolQ & 64.92 & 66.48 & 64.43 & 62.97 & 63.12 & 64.37 & \bf 65.08 & \bf 67.71 & \bf 66.64 & \bf 66.33 & \bf 66.36 & \bf 66.43 \\
\bf HelS (10) & \bf 72.03 & \bf 72.05 & \bf 72.06 & \bf 72.00 & \bf 72.12 & \bf 72.05 & 69.22 & 69.24 & 69.02 & 69.04 & 69.17 & 69.14 \\
\bf LAMB & 66.18 & 66.31 & 66.19 & 66.43 & 67.01 & 66.41 & \bf 66.91 & \bf 67.13 & \bf 68.08 & \bf 67.18 & \bf 67.77 & \bf 67.41 \\
\bf LogiQA & 26.27 & 26.27 & \bf 27.65 & 29.95 & 27.50 & 27.50 & \bf 28.73 & \bf 27.96 & 27.19 & \bf 30.11 & \bf 28.57 & \bf 28.51 \\
\bf MMLU (5) & 25.70 & 24.81 & 25.21 & 25.22 & 25.65 & 25.32 & \bf 26.99 & \bf 26.75 & \bf 27.00 & \bf 26.81 & \bf 27.01 & \bf 26.91 \\
\bf NQ (5) & 13.49 & 13.60 & 13.71 & 13.19 & 13.46 & 13.50 & \bf 15.82 & \bf 16.23 & \bf 15.96 & \bf 15.84 & \bf 16.09 & \bf 15.99 \\
\bf PIQA & \bf 76.71 & \bf 76.88 & \bf 76.17 & \bf 75.52 & \bf 75.95 & \bf 76.27 & 75.19 & 74.21 & 75.19 & 74.86 & 74.70 & 74.83 \\
\bf SciQ & \bf 90.10 & \bf 90.30 & 91.70 & 88.10 & 91.50 & 90.34 & 89.80 & 89.40 & \bf 92.70 & \bf 89.10 & \bf 91.80 & \bf 90.56 \\
\bf SQuAD & \bf 49.17 & \bf 44.33 & \bf 50.18 & \bf 51.86 & \bf 37.49 & \bf 46.60 & 44.69 & 37.94 & 47.94 & 44.93 & 30.81 & 41.25 \\
\bf TriQA & 40.14 & 40.11 & 40.06 & 39.72 & 40.43 & 40.09 & \bf 43.33 & \bf 41.84 & \bf 43.70 & \bf 43.02 & \bf 43.44 & \bf 43.07 \\
\bf TruthQA & 28.41 & 30.40 & 29.74 & \bf 29.87 & \bf 30.05 & 29.71 & \bf 30.13 & \bf 31.03 & \bf 30.06 & 29.80 & 29.61 & \bf 30.10 \\
\bf WinoG & 63.38 & 64.17 & 63.77 & 65.04 & 64.64 & 64.20 & \bf 64.72 & \bf 64.64 & \bf 65.59 & \bf 66.54 & \bf 65.04 & \bf 65.29 \\
\bf WSC & 36.54 & 37.50 & 37.50 & 37.50 & 36.54 & 37.12 & \bf 46.15 & \bf 57.69 & \bf 43.27 & \bf 63.46 & \bf 51.92 & \bf 52.31 \\ \midrule
\bf Average & 50.72 & 50.69 & 51.08 & 50.75 & 50.20 & 50.69 & \bf 51.63 & \bf 51.69 & \bf 52.00 & \bf 52.82 & \bf 51.39 & \bf 51.89 \\
\bottomrule
\end{tabular}
\caption{Performance comparison between ReSheared 2.7B and DRPruning 2.7B across five different prompts. Abbreviations follow the definitions in Table~\ref{tab:comparison13}.}
\label{tab:comparison27}
\end{table*}

\begin{table*}[!t]
\small
\centering
\begin{tabular}{p{1.45cm}p{13.7cm}}
\toprule
\multirow{5}{*}{\parbox{1.6cm}{\textbf{ARCC, ARCE, BoolQ, NQ, PIQA, SciQ, TriQA}}} & [\texttt{Passage}]. Question: [\texttt{Question}]. Answer: \\ \dashline{2-2}
 & [\texttt{Passage}]. Q: [\texttt{Question}]. A: \\ \dashline{2-2}
 & [\texttt{Passage}]. Answer the question [\texttt{Question}]. Answer: \\ \dashline{2-2}
 & [\texttt{Passage}]. Please respond to the following question: [\texttt{Question}]. Response: \\ \dashline{2-2}
 & [\texttt{Passage}]. Please answer the following: [\texttt{Question}]. Answer: \\ \midrule
 \multirow{5}{*}{\parbox{1.6cm}{\textbf{HelS, LAMB, WinoG}}} & [\texttt{Sentence}]. \\ \dashline{2-2}
 & Continue the narrative below: [\texttt{Sentence}]. \\ \dashline{2-2}
 & Provide a logical continuation for the text below: [\texttt{Sentence}]. \\ \dashline{2-2}
 & Extend the following scenario: [\texttt{Sentence}]. \\ \dashline{2-2}
 & Please carry on with the next part of the story: [\texttt{Sentence}]. \\ \midrule
 \multirow{10}{*}{\parbox{3cm}{\textbf{LogiQA}}} &  Passage: [\texttt{Passage}]. Question: [\texttt{Question}]. Choices: A. [\texttt{Choice1}]. B. [\texttt{Choice2}]. C. [\texttt{Choice3}]. D. [\texttt{Choice4}]. Answer: \\ \dashline{2-2}
 & Here is a passage: [\texttt{Passage}]. Based on the above, answer the following question: [\texttt{Question}]. Select the correct option: A. [\texttt{Choice1}]. B. [\texttt{Choice2}]. C. [\texttt{Choice3}]. D. [\texttt{Choice4}]. Your answer: \\ \dashline{2-2}
 & **Passage:** [\texttt{Passage}]. **Question:** [\texttt{Question}]. **Choices:** - A. [\texttt{Choice1}]. - B. [\texttt{Choice2}]. - C. [\texttt{Choice3}]. - D. [\texttt{Choice4}]. **Answer:** \\ \dashline{2-2}
 & Passage: [\texttt{Passage}]. \#\#\# Question: [\texttt{Question}]. \#\#\#\# Options: A) [\texttt{Choice1}]. B) [\texttt{Choice2}]. C) [\texttt{Choice3}]. D) [\texttt{Choice4}]. \#\#\# Answer: \\ \dashline{2-2}
 & You are given the following passage: [\texttt{Passage}]. Answer the question based on the passage: [\texttt{Question}]. Select one of the following options: A) [\texttt{Choice1}]. B) [\texttt{Choice2}]. C) [\texttt{Choice3}]. D) [\texttt{Choice4}]. Your Answer: \\ \midrule
 \multirow{9}{*}{\parbox{3cm}{\textbf{MMLU}}} & Q: [\texttt{Question}]. (A) [\texttt{Choice1}] (B) [\texttt{Choice2}] (C) [\texttt{Choice3}] (D) [\texttt{Choice4}] A: \\ \dashline{2-2}
 & Please provide the correct answer to the math problem below: [\texttt{Question}]. A. [\texttt{Choice1}]. B. [\texttt{Choice2}]. C. [\texttt{Choice3}]. D. [\texttt{Choice4}]. Answer: \\ \dashline{2-2}
 & Determine the solution to the following: [\texttt{Question}]. A. [\texttt{Choice1}]. B. [\texttt{Choice2}]. C. [\texttt{Choice3}]. D. [\texttt{Choice4}]. Answer: \\ \dashline{2-2}
 & What is the correct answer to the following question? [\texttt{Question}]. A. [\texttt{Choice1}]. B. [\texttt{Choice2}]. C. [\texttt{Choice3}]. D. [\texttt{Choice4}]. Answer: \\ \dashline{2-2}
 & What is the solution to this math problem? [\texttt{Question}]. Options: A) [\texttt{Choice1}]. B) [\texttt{Choice2}]. C) [\texttt{Choice3}]. D) [\texttt{Choice4}]. Answer: \\ \midrule
 \multirow{5}{*}{\parbox{3cm}{\textbf{SQuAD}}} & Title: [\texttt{Title}]. Background: [\texttt{Context}]. Question: [\texttt{Question}]. Answer: \\ \dashline{2-2}
 & Context: [\texttt{Context}]. Question: [\texttt{Question}]. Answer: \\ \dashline{2-2}
 & Given the following text: [\texttt{Context}]. Answer the question below: [\texttt{Question}]. Answer: \\ \dashline{2-2}
 & Information: [\texttt{Title}]. [\texttt{Context}]. Please answer the following: [\texttt{Question}]. Answer: \\ \dashline{2-2}
 & Background Information: [\texttt{Context}]. Please address the following question: [\texttt{Question}]. Answer: \\ \midrule
 \multirow{5}{*}{\parbox{3cm}{\textbf{TruthQA}}} & [\texttt{TruthQA Few Shot}]. Q: [\texttt{Question}]. A: \\ \dashline{2-2}
 & [\texttt{TruthQA Few Shot}]. What is the answer to this question? [\texttt{Question}]. A: \\ \dashline{2-2}
 & [\texttt{TruthQA Few Shot}]. Question: [\texttt{Question}]. Provide your answer: \\ \dashline{2-2}
 & [\texttt{TruthQA Few Shot}]. Q: [\texttt{Question}]. Please provide the answer (A): \\ \dashline{2-2}
 & [\texttt{TruthQA Few Shot}]. Provide an answer to the following question: [\texttt{Question}]. Answer: \\ \midrule
 \multirow{9}{*}{\parbox{3cm}{\textbf{WSC}}} & Passage: [\texttt{Passage}]. Question: In the passage above, does the pronoun "*[\texttt{Pronoun}]*" refer to "*[\texttt{Noun}]*"? Answer: \\ \dashline{2-2}
 & Analyze the following text: [\texttt{Passage}]. Question: Is the pronoun "*[\texttt{Pronoun}]*" referring to "*[\texttt{Noun}]*"? Answer: \\ \dashline{2-2}
 & Examine the following passage: [\texttt{Passage}]. Question: In this passage, does the pronoun "*[\texttt{Pronoun}]*" refer to "*[\texttt{Noun}]*"? Answer: \\ \dashline{2-2}
 & Passage Analysis: [\texttt{Passage}]. Question: Does the pronoun "*[\texttt{Pronoun}]*" in the passage refer to "*[\texttt{Noun}]*"? Answer: \\ \dashline{2-2}
 & Analyze the text below: [\texttt{Passage}]. Question: Is the pronoun "*[\texttt{Pronoun}]*" referring to "*[\texttt{Noun}]*"? Answer: \\
\bottomrule
\end{tabular}
\caption{Prompts used for significance testing. For each task, we designed five prompts.}
\label{tab:comparisonPrompt}
\end{table*}

\subsection{Robustness Verification}
\label{apx:ttest}

First, all comparisons in our experiments are mainly based on \textbf{DRPruning} and \textbf{ReSheared}, rather than the official open-source version of Sheared LLaMA. This is because, under the 2.7B configuration, we were unable to reproduce the results using RedPajama or the filtered SlimPajama dataset as the continued pretrained dataset. However, for the 1.3B model, our reproduced version achieved performance surpassing Sheared LLaMA. Therefore, to ensure a fair comparison, we conducted most comparisons against ReSheared.

As shown in Table~\ref{tab:main}, our method demonstrates relatively small improvements over ReSheared in downstream evaluations. To address this issue, we provide the following analysis. First, our results consistently outperform ReSheared across various metrics, including PPL, downstream task performance for both pruned and continued pretrained models, domain-specific evaluation, and win rate after instruction tuning. These consistent and stable improvements across multiple dimensions provide solid evidence of the effectiveness of our approach.

To further demonstrate the robustness of our method, we conduct significance testing. Specifically, we design five distinct prompts for each task to test its resilience to input perturbations, and conduct paired t-tests between ReSheared and DRPruning. The results under the 1.3B and 2.7B configurations are presented in Table~\ref{tab:comparison13} and \ref{tab:comparison27}, respectively. For the 1.3B model, the t-statistic is 2.0318 with a p-value of 0.0458, while for the 2.7B model, the t-statistic is 2.1962 with a p-value of 0.0312. When combined, the overall t-statistic reaches 2.9922 with a p-value of 0.0032. These results provide strong evidence of the statistical significance of our method, with a p-value below 0.05. The prompts used are given in Table~\ref{tab:comparisonPrompt}.

\begin{table}[!ht]
\small
\centering
\scalebox{0.89}{
\begin{tabular}{lccccc}
\toprule
\multirow{2}[2]{*}{\bf Method} & \multicolumn{2}{c}{\bf Pruning} &  \multicolumn{3}{c}{\bf Cont. PT} \\ \cmidrule(lr){2-3}\cmidrule(lr){4-6}
~ & \bf PPL $\downarrow$ & \bf Time $\downarrow$ & \bf PPL $\downarrow$ & \bf Task $\uparrow$ & \bf Time $\downarrow$ \\ \midrule
\bf ReSheared & 20.07 & 43.96 & 8.37 & 36.33 & 225.42 \\
\bf DRPruning & \bf 16.88 & \bf 43.74 & \bf 7.68 & \bf 36.49 & \bf 221.85 \\
\bottomrule
\end{tabular}}
\caption{\label{tab:efficiency}
PPL, training time (in hours), and downstream task performance (Task) of 0.5B pruned models.}
\end{table}

\subsection{Efficiency Discussion}
\label{apx:efficiency}
DRPruning focuses solely on data distribution without introducing additional GPU computations. The only extra cost stems from data ratio calculation, which is entirely handled on CPU. During continued pretraining, each update takes 39.02s, while pruning with an additional parameter increases it to 99.52s. Over the full training process, pruning adds 1.8 hours, and continued pretraining adds 1.3 hours. This accounts for a 1.3\% increase in training time for the 1.3B model and 0.7\% for the 2.7B model.

To eliminate extra computation overhead, we implemented parallel data ratio calculation, ensuring training remains uninterrupted. This introduces a one- to two-step update delay, which does not affect performance. To prove this, a small-scale experiment, following the main setup, is conducted with a 0.5B target model for 24k steps in continued pretraining.

Results are in Table~\ref{tab:efficiency}. On four NVIDIA A800 80GB GPUs, our method requires less training time after parallelization. However, before parallelization, pruning takes 44.15 hours, which is longer than ReSheared. PPL is 17.01 and 16.88 before and after parallelization, respectively, demonstrating that parallelization improves efficiency without compromising performance.

Additionally, our method maintains a lower PPL, outperforming ReSheared. However, improvements in downstream tasks are marginal, with performance on many tasks approaching or even falling below random guessing. Given the extremely high PPL after pruning (16.88 for 0.5B, 9.83 for 1.3B, 7.40 for 2.7B), we conclude that pruning from 7B to 0.5B leads to a performance collapse, making effective recovery challenging.

\subsection{Analysis of More Fine-Grained Domains}
\label{apx:finegrained}

DRPruning demonstrates superior performance in fine-grained domain segmentation. To substantiate this claim, we conduct further analysis on pruning experiments from Llama-2-7B to 1.3B. We then perform a more detailed domain segmentation by dividing CC into 10 parts and C4 into 3 parts, ensuring that each domain accounts for approximately 5\% of the total data, indicated as fine-grained (\textbf{Fine}). Specifically, we use the all-MiniLM-L6-v2~\citep{DBLP:conf/nips/WangW0B0020} model to encode the input sentences, followed by k-means clustering to segment the data into smaller domains. For each fine-grained domain, we retrain 100 samples as the validation set. As for the test set, we still use the one provided by Sheared Llama, with 500 samples each for the coarse-grained domains. We conduct pruning for 1600 steps, with other experimental settings aligned with those in our paper. We compare this with the method before segmentation, indicated as coarse-grained (\textbf{Coarse}).

Results are in Table~\ref{tab:finegrained}, where we report the validation set cross-entropy (CE) for the segmented domains, where the category names are summarized by Deepseek R1 \citep{deepseekai2025deepseekr1incentivizingreasoningcapability} from 100 texts within each domain. Results demonstrate that fine-grained segmentation can accelerate the convergence, especially in the early stage of training, and obtain better performance across these domains. After more extensive training, the advantage is still maintained in the majority of domains.

We also report the results on the test set, as shown in Table~\ref{tab:coarse_test}, which lead to similar conclusions as those on the validation set across all domains. Furthermore, we migrate the fine-grained method to ReSheared and find that the performance is not satisfactory. The reason could be that, with more fine-grained domains, the hyperparameter settings for reference loss and data ratio become more complex, making our dynamic hyperparameter facility method more critical.

\begin{table}[!t]
\small
\centering
\scalebox{0.97}{\begin{tabular}{lcccc}
\toprule
\multirow{2}[2]{*}{\bf Domain} & \multicolumn{2}{c}{\textbf{800 Steps}} & \multicolumn{2}{c}{\textbf{1600 Steps}} \\
\cmidrule(lr){2-3} \cmidrule(lr){4-5}
& \textbf{Coarse} & \textbf{Fine} & \textbf{Coarse} & \textbf{Fine} \\
\midrule
\textbf{CC-music} & 3.008 & \textbf{2.974} & 2.844 & \textbf{2.835} \\
\textbf{CC-technology} & 2.912 & \textbf{2.877} & 2.753 & \textbf{2.736} \\
\textbf{CC-sports} & 3.013 & \textbf{2.969} & 2.851 & \textbf{2.827} \\
\textbf{CC-environment} & 2.707 & \textbf{2.698} & 2.568 & \textbf{2.567} \\
\textbf{CC-medical} & 2.789 & \textbf{2.734} & 2.629 & \textbf{2.590} \\
\textbf{CC-corporate} & 2.743 & \textbf{2.691} & 2.581 & \textbf{2.553} \\
\textbf{CC-entertainment} & 3.070 & \textbf{3.057} & \textbf{2.899} & 2.911 \\
\textbf{CC-politics} & 3.033 & \textbf{3.001} & 2.865 & \textbf{2.858} \\
\textbf{CC-legal} & 3.065 & \textbf{3.033} & 2.892 & \textbf{2.891} \\
\textbf{CC-culture} & 3.099 & \textbf{3.067} & 2.936 & \textbf{2.927} \\
\textbf{C4-forum} & 3.112 & \textbf{3.086} & 2.955 & \textbf{2.946} \\
\textbf{C4-business} & 3.139 & \textbf{3.107} & 2.982 & \textbf{2.969} \\
\textbf{C4-lifestyle} & 3.178 & \textbf{3.157} & 3.019 & \textbf{3.013} \\
\midrule
\textbf{Average} & 2.990 & \textbf{2.958} & 2.829 & \textbf{2.817} \\
\bottomrule
\end{tabular}}
\caption{Cross-entropy of using more fine-grained domains on the validation set. ``Coarse'' and ``Fine'' means using coarse-grained or fine-grained domain split. ``800 Steps'' and ``1600 Steps'' means the total training step for pruning.}
\label{tab:finegrained}
\end{table}

\begin{table*}[h!]
\small
\centering
\begin{tabular}{lcccccc}
\toprule
\multirow{2}[3]{*}{\bf Domain} & \multicolumn{3}{c}{\textbf{800 Steps}} & \multicolumn{3}{c}{\textbf{1600 Steps}} \\
\cmidrule(lr){2-4} \cmidrule(lr){5-7}
& \makecell{\textbf{ReSheared} \\ \textbf{Fine}} & \makecell{\textbf{DRPruning} \\ \textbf{Coarse}} & \makecell{\textbf{DRPruning} \\ \textbf{Fine}} & \makecell{\textbf{ReSheared} \\ \textbf{Fine}} & \makecell{\textbf{DRPruning} \\ \textbf{Coarse}} & \makecell{\textbf{DRPruning} \\ \textbf{Fine}} \\
\midrule
\textbf{CC} & 3.373 & 2.886 & \textbf{2.856} & 3.196 & 2.723 & \textbf{2.714} \\
\textbf{C4} & 3.676 & 3.128 & \textbf{3.102} & 3.512 & 2.969 & \textbf{2.959} \\
\textbf{GitHub} & 2.189 & 1.446 & \textbf{1.421} & 2.059 & 1.286 & \textbf{1.277} \\
\textbf{Book} & 3.594 & 3.131 & \textbf{3.117} & 3.383 & \textbf{2.946} & 2.956 \\
\textbf{Wiki} & 2.978 & 3.550 & \textbf{3.491} & \textbf{2.404} & 3.261 & 3.268 \\
\textbf{ArXiv} & 3.184 & 2.185 & \textbf{2.170} & 3.034 & 2.022 & \textbf{2.016} \\
\textbf{StackExchange} & \textbf{3.278} & 2.496 & 2.473 & 3.157 & 2.321 & \textbf{2.313} \\
\midrule
\textbf{Average} & 3.180 & 2.685 & \textbf{2.658} & 2.963 & 2.501 & \textbf{2.497} \\
\bottomrule
\end{tabular}
\caption{\label{tab:coarse_test}
Cross-entropy of using more fine-grained domains on the test set. All abbreviations used here align with Table~\ref{tab:domain} and Table~\ref{tab:finegrained}.}
\end{table*}

\begin{table}[!t]
\small
\centering
\begin{tabular}{llcc}
\toprule
\textbf{Language} & \textbf{Data}  & \textbf{ReSheared} & \textbf{DRPruning} \\
\midrule
\textbf{EN-ZH}   & 39.9k & 25.4 & \textbf{27.4} \\
\textbf{ZH-EN}   & 39.9k & 21.9 & \textbf{23.1} \\
\textbf{EN-DE}   & 39.3k & 21.5 & \textbf{26.2} \\
\textbf{DE-EN}   & 39.3k & 32.8 & \textbf{33.1} \\
\textbf{EN-RU}   & 34.0k & 20.9 & \textbf{24.1} \\
\textbf{RU-EN}   & 34.0k & 26.7 & \textbf{27.2} \\
\textbf{EN-AR}   & 17.4k & 9.4 & \textbf{9.6} \\
\textbf{AR-EN}   & 17.4k & 21.0 & \textbf{24.5} \\
\textbf{EN-ID}   & 2.9k  & 11.5 & \textbf{20.9} \\
\textbf{ID-EN}   & 2.9k  & \textbf{27.4} & 26.8 \\
\textbf{EN-HI}   & 2.2k  & \textbf{5.6} & 1.5 \\
\textbf{HI-EN}   & 2.2k  & 7.5 & \textbf{14.2} \\
\textbf{EN-JA}   & 0.8k  & 5.1 & \textbf{8.5} \\
\textbf{JA-EN}   & 0.8k  & 15.0 & \textbf{18.6} \\ \midrule
\textbf{Average} &       & 18.0 & \textbf{20.4} \\
\bottomrule
\end{tabular}
\caption{\label{tab:mt_bleu}
BLEU Scores for instruction tuning analysis on machine translation. ``Language'' indicates the source and target language pair, connected by a hyphen. ``Data'' refers to the training data volume.}
\end{table}

\begin{table}[!t]
\small
\centering
\scalebox{0.95}{
\begin{tabular}{lccc}
\toprule
\textbf{Language} & \textbf{Default Ratio} & \textbf{ReSheared} & \textbf{DRPruning} \\
\midrule
\textbf{EN-ID}   & 1.05   & 0.42 & 2.03 \\
\textbf{ID-EN}   & 1.05   & 0.11 & 1.27 \\
\textbf{EN-HI}   & 0.81   & 72.14 & 0.93 \\
\textbf{HI-EN}   & 0.81   & 0.48 & 1.05 \\
\textbf{EN-JA}   & 0.31   & 0.47 & 2.14 \\
\textbf{JA-EN}   & 0.31   & 0.05 & 0.44 \\
\bottomrule
\end{tabular}}
\caption{\label{tab:mt_data_ratios}
Data ratios for instruction tuning analysis on machine translation. ``Default Ratio'' denotes the percentage of the total training data comprised by each domain's data, as determined by its original volume.}
\end{table}

\subsection{Analysis on Instruction Tuning}
\label{apx:instuning}

DRPruning demonstrates efficacy not only under pruning and continued pretraining but also under instruction tuning. To verify this, we conduct experiments on machine translation tasks. Specifically, for training data, we employ the News-Commentary~\citep{DBLP:conf/wmt/KocmiABBDFFFGGH23} dataset, and perform downsampling. For domains with larger data volumes, we sample 10\% of the data, while for those with smaller volumes, we sample 50\%, using the sampled data ratios as the default. For validation and testing, we select the FLORES-200 dev and devtest sets \citep{nllb2022}, respectively. Following \citet{ParroTTranslating_JHW+23}, we adopt 33 prompts during training, and mask the instruction, training solely on the response. Regarding training, we utilize Qwen2-1.5B as the base model, with a batch size of 128 over 2k steps and a learning rate of 2e-5. The reference loss is computed based on Qwen2-7B, and other settings remain consistent with the continued pretraining described in the main paper.

The experimental results are shown in Table~\ref{tab:mt_bleu}, reporting BLEU scores across all language directions. Our method demonstrates significant improvements over ReSheared. To analyze the reasons, we report the data ratios used during training in Table~\ref{tab:mt_data_ratios}. When the performance in a certain domain consistently falls below expectations, ReSheared leads to distribution collapse in scenarios with distribution shift, resulting in excessive training epochs on a small amount of data. However, our method, by dynamically adjusting expectations (dynamic reference loss) and data ratio constraints (DRO + dynamic reference data ratio), robustly allocates more weight to low-resource languages without significantly deviating from the preset data distribution.

\end{document}